\documentclass[lettersize,journal]{IEEEtran}

\usepackage{amsmath,amsfonts,amssymb}
\usepackage{graphicx}
\usepackage{url}
\usepackage{array}
\usepackage{booktabs}
\usepackage{tabularx}
\usepackage{multicol}
\usepackage{microtype}
\usepackage[caption=false,font=normalsize,labelfont=sf,textfont=sf]{subfig}
\usepackage[table]{xcolor}
\usepackage{algorithm}
\usepackage{algpseudocode}
\usepackage{tcolorbox}
\tcbuselibrary{breakable,skins}
\usepackage{enumitem}
\usepackage[T1]{fontenc}
\usepackage[utf8]{inputenc}
\usepackage{pifont}
\usepackage{soul}
\usepackage[normalem]{ulem}
\usepackage{listings}
\usepackage{balance}
\usepackage{amsthm}
\usepackage[hidelinks]{hyperref}
\newtheorem{proposition}{Proposition}
\tcbset{before skip=6pt, after skip=10pt}

\newtcolorbox{appendixpromptbox}[1]{
  enhanced, breakable,
  colback=black!2,
  colframe=blue!60!black,
  colbacktitle=blue!60!black,
  coltitle=white,
  fonttitle=\bfseries\large,
  title={#1},
  arc=3mm, boxrule=1.0pt,
  left=4mm, right=4mm, top=3mm, bottom=3mm
}

\definecolor{LightGray}{gray}{0.92}
\definecolor{MidGray}{gray}{0.85}

\definecolor{plancolor}{RGB}{147,112,219}
\definecolor{memcolor}{RGB}{106,168,79}
\definecolor{retrievecolor}{RGB}{230,145,56}
\definecolor{inspectcolor}{RGB}{234,67,53}
\definecolor{solvecolor}{RGB}{66,133,244}

\definecolor{stepheader}{RGB}{230,230,250}
\definecolor{successgreen}{RGB}{0,128,0}
\definecolor{failred}{RGB}{220,20,60}
\definecolor{warningorange}{RGB}{255,140,0}

\begin{document}


\title{
    \textbf{\textsc{PRISMA}}: A Schema-Coordinated Multi-Agent Framework for Open-Domain Multi-Hop Question Answering
}

\author{Yu Liu,~Wenxiao Zhang,~Cong~Cao,~Fangfang~Yuan,~Hao Peng,~Yanbing~Liu,~Jin~B.~Hong,\\~Zhiyuan~Ma,~\IEEEmembership{Member,~IEEE},
and~Bowen Zhou,~\IEEEmembership{Fellow,~IEEE}%
\thanks{This work was supported in part by the National Key Research and Development Program of China under Grant 2023YFC3303800, and in part by the National Natural Science Foundation of China under Grants U25B2029, 62322202, and 62406161.}%
\thanks{\copyright\ 2026 IEEE. Personal use of this material is permitted. IEEE permission is required for any other use in any medium, including promotional reprints, new collective works, resale, redistribution to servers or mailing lists, and reuse of any copyrighted component in other works.}%
\thanks{\textit{(Yu Liu and Wenxiao Zhang contributed equally to this work.)} \textit{(Corresponding authors: Cong Cao and Zhiyuan Ma.)}}%
\thanks{Yu Liu and Yanbing Liu are with the Institute of Information Engineering,
Chinese Academy of Sciences, Beijing 100085, China, and also with the School of Cyber Security, University of Chinese Academy of Sciences, Beijing 100049, China (e-mail: \{liuyu, liuyanbing\}@iie.ac.cn).}%
\thanks{Wenxiao Zhang and Jin B. Hong are with the Department of Computer Science and Software Engineering, University of Western Australia, Perth, Australia, 6009 (e-mail: wenxiao.zhang@research.uwa.edu.au, jin.hong@uwa.edu.au).}%
\thanks{Cong Cao and Fangfang Yuan are with the Institute of Information Engineering,
Chinese Academy of Sciences, Beijing 100085, China (e-mail: \{caocong, yuanfangfang\}@iie.ac.cn).}%
\thanks{Hao Peng is with State Key Laboratory of Software Development Environment, Beihang University, Beijing 100191, China (e-mail: penghao@buaa.edu.cn).}%
\thanks{Zhiyuan Ma is with Huazhong University of Science and Technology, Wuhan 430074, China (e-mail: mzyth@hust.edu.cn).}%
\thanks{Bowen Zhou is with the Department of Electronic Engineering, Tsinghua University, Beijing 100084, China (e-mail: zhoubowen@tsinghua.edu.cn).}%
}
\maketitle


\begin{abstract}
Constructing reliable evidence chains across massive corpora for open-domain multi-hop question answering remains a core challenge in knowledge-intensive information retrieval. End-to-end approaches that retrieve and reason in a single pass struggle with multi-hop queries, as a single retrieval step cannot simultaneously locate all bridge passages required for chained reasoning. This necessitates multi-step pipelines, yet these face two bottlenecks:
\textbf{\emph{1) Evidence Chain Fragility.}} Retrieval errors at early hops cascade and amplify through subsequent steps without reasoning-guided planning.
\textbf{\emph{2) Cross-Module Attribution.}} Multi-step trajectory training cannot attribute errors to specific pipeline stages, limiting generalization. To address these issues, we propose \textbf{\textsc{PRISMA}}, a multi-agent framework with a \underline{P}lan-\underline{R}etrieve-\underline{I}nspect-\underline{S}olve-\underline{M}emoize \underline{A}rchitecture that reflects expert information seeking behavior. PRISMA elevates structured generation from an intra-model constraint to an inter-agent communication schema, where each agent's output simultaneously enforces its own reasoning discipline and defines the input interface for downstream agents. An Inspector acts as an in-loop verifier, auditing the evidence chain at each hop and triggering targeted recovery. We adopt a decoupled training strategy with agent-specific rewards: Stage~I specializes the Planner and Solver as task experts; Stage~II freezes the trained pipeline and trains the Inspector on trajectory-conditioned verification, as stable execution traces are essential for learning meaningful error patterns. A recovery-gain analysis quantifies the net contribution of the verification stage. Across all ten benchmarks, \textsc{PRISMA} consistently outperforms all trained baselines and delivers overall performance competitive with larger closed-source API models, supporting efficient real-world deployment. The code is available at \url{https://github.com/prisma-research/PRISMA}.
\end{abstract}

\begin{IEEEkeywords}
Multi-agent systems, retrieval-augmented generation, multi-hop question answering, inter-agent communication schema, evidence chain construction, reinforcement learning.
\end{IEEEkeywords}


\begin{figure*}[!t]
 \centering
 \includegraphics[width=\textwidth]{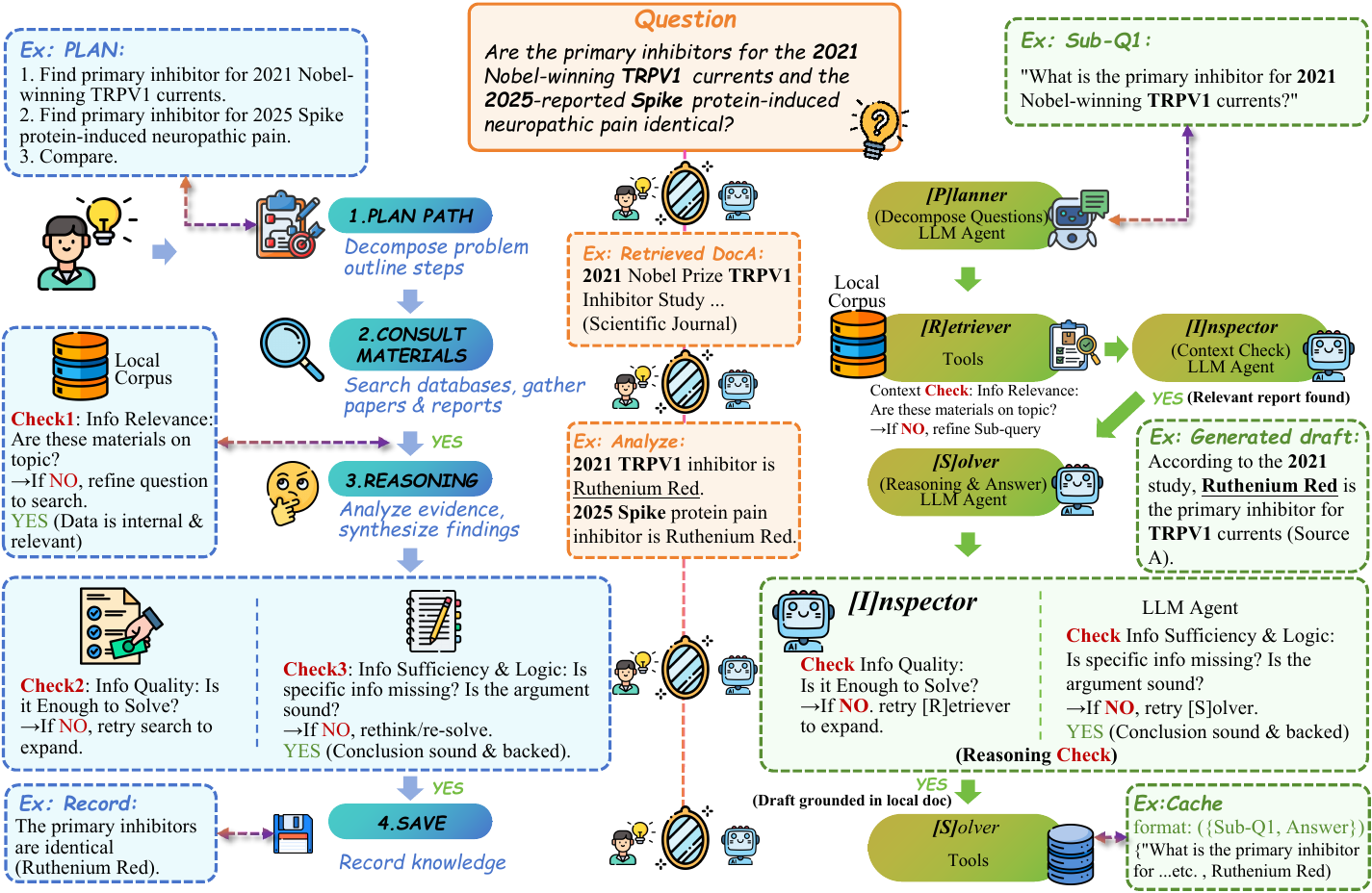}
\caption{%
\textbf{\textsc{PRISMA}'s agent pipeline reflects the iterative search-verify-synthesize process in expert information seeking.}
\textbf{Left:} Researchers tackle complex questions (e.g., comparing \textbf{protein inhibitors} from the \textbf{2021 Nobel-winning TRPV1} and \textbf{2025 Spike protein} studies) by (1) \emph{Planning} the problem; (2) \emph{Consulting materials}; (3) \emph{Reasoning} by analyzing evidence and checking it against the question; (4) \emph{Answering} the synthesis question; and (5) \emph{Noting} key findings in a notebook for future reference.
\textbf{Right:} \textbf{\textsc{PRISMA}} replicates this with agents: \textbf{Planner} decomposes questions, \textbf{Retriever} gathers documents, \textbf{Inspector} validates relevance and reasoning, \textbf{Solver} synthesizes answers, and \textbf{Memoize} records solved subquestions. Targeted retrieval expansion and chain-of-reasoning feedback complete the pipeline.
}
 \label{fig:motivation}
\end{figure*}

\section{Introduction}
\IEEEPARstart{A}{s} a typical dynamic interacting system, open-domain multi-hop question answering (QA) is a knowledge-intensive information retrieval task that requires constructing reliable evidence chains across massive corpora. A complex query like \textit{``Are the primary inhibitors for 2021 Nobel-winning \textbf{TRPV1} currents and 2025-reported \textbf{Spike} protein-induced neuropathic pain identical?''} cannot be answered by a single retrieval step; it demands three tightly coupled capabilities:
\textbf{\textit{First}}, planning a dependency-aware chain of subquestions that surfaces intermediate \emph{bridge entities}.
\textbf{\textit{Second}}, retrieving the right supporting passages for each hop from millions of candidates.
\textbf{\textit{Third}}, reasoning over the collected evidence to produce a grounded final answer.
When any one capability fails, the entire evidence chain breaks.

Researchers have explored various paradigms to improve multi-hop QA.
\textit{\textbf{(1) Single-Step RAG:}} Conventional single-step RAG~\cite{lewis2020retrieval} retrieves once using the full query. When intermediate entities are not explicit in the original question, this approach may fail to retrieve all supporting passages needed for multi-hop reasoning.
\textit{\textbf{(2) Iterative Multi-Step RAG:}} ReAct~\cite{yao2023react}, IRCoT~\cite{trivedi2023interleaving}, RAG-Star~\cite{jiang2025rag}, and Search-o1~\cite{li2025search} interleave retrieval with reasoning to progressively gather evidence. However, without explicit planning, retrieval errors at early hops cascade and amplify through subsequent steps, causing bridge answer failures.
\textit{\textbf{(3) Training-Based and Self-Reflective Approaches:}} Self-RAG~\cite{asai2024selfrag} uses reflection tokens to guide retrieval and generation, while Retrobust~\cite{yoran2024robust} improves robustness to irrelevant retrieved context through supervised fine-tuning. RAG-DDR~\cite{li2025ragddr}, BGM~\cite{ke2024bridging}, and ARENA~\cite{ren2025arena} optimize RAG components, retriever--generator alignment, and evidence traceability, respectively. This motivates explicit coordination of component interfaces to support cross-agent error attribution and targeted recovery.
\textit{\textbf{(4) Multi-Agent Pipelines:}} Recent works like MMOA~\cite{chen2025mmoa}, TIRESRAG-R1~\cite{he2025tiresrag}, and QAgent~\cite{jiang2025qagent} distribute subtasks across specialized agents. However, they lack effective inter-agent communication protocols and in-loop verification mechanisms, leaving retrieval errors undetected. Multi-step trajectory training further suffers from cross-module attribution difficulties, limiting generalization.
These limitations share a common root: the absence of a schema-defined coordination mechanism that enables both modular error attribution and targeted recovery within the retrieval pipeline.

To mitigate these challenges, we introduce \textbf{\textsc{PRISMA}} (\textbf{P}lan-\textbf{R}etrieve-\textbf{I}nspect-\textbf{S}olve-\textbf{M}emoize \textbf{A}rchitecture), a multi-agent framework that reflects the iterative search-verify-synthesize process observed in expert information seeking (Fig.~\ref{fig:motivation}).
Relative to iterative and verification-based RAG systems, which interleave retrieval and refinement within an evolving trajectory~\cite{trivedi2023interleaving,jiang2025rag,li2025search}, and modular multi-agent approaches, which emphasize role specialization, reflection, or interactive query understanding~\cite{chen2025mmoa,he2025tiresrag,jiang2025qagent}, PRISMA jointly couples schema-defined inter-agent contracts, dual-phase in-loop verification, diagnosis-specific recovery, and responsibility-aligned agent training.
\textbf{\textit{(1) Architecture:}} PRISMA elevates structured generation from an intra-model formatting constraint to an inter-agent communication schema: each agent's structured output simultaneously enforces its own reasoning discipline and defines the input interface for downstream agents. The \textbf{Planner} decomposes a query into dependency-aware subquestions to drive fine-grained retrieval, while the \textbf{Inspector} serves as an in-loop verifier that diagnoses whether retrieved evidence is sufficient (Context Inspector) and whether the Solver's answer is faithfully grounded (Reasoning Inspector), triggering targeted recovery including retrieval expansion, subquestion rewriting, or answer retry.
\textbf{\textit{(2) Training:}} We adopt a decoupled training strategy with agent-specific rewards optimized via GRPO~\cite{shao2024deepseekmath}. Stage~I specializes the Planner and Solver as task experts. Stage~II freezes the trained pipeline and trains the Inspector on trajectory-conditioned verification, as stable execution traces are essential for learning meaningful error patterns. Our contributions are \textbf{threefold}:
\begin{itemize}[leftmargin=*]

\item \textbf{Schema-Defined Multi-Agent Architecture for Evidence Chain Construction.}
 We introduce PRISMA, which coordinates a Planner, Solver, and dual-phase Inspector through a structured inter-agent communication schema, enabling modular error attribution and targeted recovery at each hop.

\item \textbf{Decoupled Training with Agent-Specific Rewards.} We propose a decoupled training strategy that mitigates the credit attribution difficulty in multi-agent pipelines. Each agent is optimized within its schema-defined responsibility boundary, with Stage I specializing individual agents and Stage II training the Inspector on a frozen pipeline for stable verification learning.

\item \textbf{Comprehensive Open-Domain Evaluation.} Across all ten benchmarks over a 21M-passage corpus, \textsc{PRISMA} consistently surpasses the reported trained baselines and delivers overall performance competitive with larger closed-source API models, with particularly consistent out-of-distribution generalization where existing RL-trained methods degrade.
\end{itemize}


\section{Preliminary}

\subsection{Evaluation Settings: Open-Domain vs. Closed-Domain}
\label{sec:eval_settings}
Following established multi-hop QA taxonomies and benchmark protocols~\cite{mavi2024multi,yang2018hotpotqa}, we distinguish the settings by whether a question-specific set of sufficient evidence is provided as task input. Corpus size and the number of passages read by the model are separate experimental choices.

\noindent\textbf{Closed-domain (reader setting).} Each question is paired with a candidate document set that contains sufficient supporting evidence and may also include distractors. Evaluation therefore focuses on evidence selection and multi-hop reasoning within the provided set, without requiring corpus-level retrieval to find the supporting documents.

\noindent\textbf{Open-domain setting.} The system receives the question without a question-specific evidence set and must retrieve relevant passages from the available corpus $\mathcal{C}$ before reasoning. This setting exposes the full pipeline to retrieval uncertainty, where missing or noisy evidence can propagate into downstream reasoning. We instantiate this setting with the full 21M-passage DPR Wikipedia corpus; dataset- and corpus-level details are provided in Section~\ref{sec:experimental_setup}.

\subsection{Problem Formulation: Open-Domain Multi-Hop QA}
\label{sec:problem_formulation}
Given a query $q$ and an open-domain corpus $\mathcal{C} = \{d_1, \ldots, d_N\}$, the task is to produce an answer $a$ and a set of supporting facts $\mathcal{F} \subset \mathcal{C}$. Multi-hop queries require $k$ reasoning steps ($k \geq 2$), where each hop depends on bridge entities discovered in previous hops. Formally, the system must: (i) decompose $q$ into a dependency-aware chain of subquestions $\{sq_1, \ldots, sq_k\}$, (ii) retrieve supporting passages $\{p_1, \ldots, p_k\}$ from $\mathcal{C}$ for each hop, and (iii) synthesize the collected evidence into a grounded answer $a$. Let $R_i$ denote successful retrieval at hop $i$. Because later-hop queries are instantiated using earlier intermediate answers, these events are generally dependent, and the complete-chain success probability follows the chain rule, $P(\bigcap_{i=1}^{k}R_i)=P(R_1)\prod_{i=2}^{k}P(R_i\mid R_1,\ldots,R_{i-1})$. Under the simplifying approximation that the initial-hop recall and all subsequent conditional recalls share a common value $r$, this probability scales as $r^k$.

\subsection{Group Relative Policy Optimization}
\label{sec:grpo_preliminary}
We adopt Group Relative Policy Optimization (GRPO)~\cite{shao2024deepseekmath} as the critic-free policy optimizer for all trainable agents. Let $\pi_\theta(\cdot \mid u)$ denote a policy parameterized by $\theta$ conditioned on input $u$. For a training input $u$, the behavior policy $\pi_{\theta_{\mathrm{old}}}$ samples $K$ candidate outputs $\{z_1, \ldots, z_K\}$ and computes rewards $\{R(u,z_i)\}_{i=1}^K$. Their group-relative advantages are
\begin{equation}
\hat{A}_i = \frac{R_i - \mu_{\mathcal{G}}}{\sigma_{\mathcal{G}} + \varepsilon_{\mathrm{n}}}, \qquad \mathcal{G} = \{R_1, \ldots, R_K\},
\label{eq:grpo_norm}
\end{equation}
where $R_i:=R(u,z_i)$, $\mu_{\mathcal{G}}$ and $\sigma_{\mathcal{G}}$ are the mean and standard deviation over $\mathcal{G}$, and $\varepsilon_{\mathrm{n}}>0$ ensures numerical stability. For token $t$ in $z_i$, define the importance ratio
\begin{equation}
\begin{aligned}
\rho_{i,t}(\theta)
&=
\frac{\pi_\theta(z_{i,t}\mid u,z_{i,<t})}
{\pi_{\theta_{\mathrm{old}}}(z_{i,t}\mid u,z_{i,<t})},\\
\bar{\rho}_{i,t}(\theta)
&=\operatorname{clip}\!\left(\rho_{i,t}(\theta),
1-\varepsilon_{\mathrm{c}},1+\varepsilon_{\mathrm{c}}\right).
\end{aligned}
\label{eq:grpo_ratio}
\end{equation}
The clipped group-relative objective is
\begin{equation}
\begin{aligned}
\mathcal{J}_{\mathrm{GRPO}}(\theta)
=\mathbb{E}\Bigg[\sum_{i=1}^{K}\sum_{t=1}^{|z_i|}w_{i,t}\Big(
&\min\!\big\{\rho_{i,t}\hat{A}_i,\bar{\rho}_{i,t}\hat{A}_i\big\}\\
&-\beta\,d_{\mathrm{KL}}^{i,t}(\pi_\theta\Vert\pi_{\mathrm{ref}})
\Big)\Bigg],
\end{aligned}
\label{eq:grpo_obj}
\end{equation}
where the expectation is over training inputs and groups sampled from $\pi_{\theta_{\mathrm{old}}}$, $\varepsilon_{\mathrm{c}}$ is the clipping coefficient, $d_{\mathrm{KL}}^{i,t}$ is the token-level KL penalty against a fixed reference policy, and $\beta\geq0$ controls its strength. The weights $w_{i,t}$ specify token-loss aggregation; our agent-specific choices are described in Section~\ref{sec:training}. Thus, all agents retain GRPO's common core---group-relative credit assignment and clipped token-level policy updates---while using agent-appropriate reward functions and regularization. This critic-free design avoids a separate value network and is particularly suitable for our decoupled multi-agent setting.


\section{PRISMA Framework}

\subsection{Schema-Defined Agent Coordination}
\label{sec:agent_schema}

PRISMA coordinates agents through structured XML schemas that define each agent's input-output interface. Table~\ref{tab:agent_schema} specifies the communication protocol across the pipeline. Each schema enforces format constraints on the producing agent while providing a parseable input structure for downstream consumers. The schema boundaries create natural responsibility partitions: the Planner owns subquestion quality, the Solver owns answer grounding, and the Inspector owns error detection accuracy. This partitioning directly enables the agent-specific reward design described in Section~\ref{sec:reward_design}.

\begin{table}[!t]
\caption{Inter-Agent Communication Schema}
\label{tab:agent_schema}
\centering
\scriptsize
\setlength{\tabcolsep}{3pt}
\begin{tabular*}{\columnwidth}{@{\extracolsep{\fill}}lll@{}}
\toprule
\textbf{Agent} & \textbf{Output Schema} & \textbf{Consumer $\rightarrow$ Action} \\
\midrule
\textbf{Planner}
  & \texttt{<reasoning>}, \texttt{<subquestions>}
  & Memoizer $\rightarrow$ Retriever \\
  & with \texttt{[ANSWER\_i]} placeholders
  & (always) \\
\midrule
\textbf{Context Inspector}
  & \texttt{<error\_stage>},
  & \textbf{Planner} (rewrite) / \\
  & \texttt{<explanation>}
  & Retriever (expand) \\
\midrule
\textbf{Solver}
  & \texttt{<reasoning>}, \texttt{<sources>},
  & \textbf{Reasoning Inspector} \\
  & \texttt{<answer>}
  & (after context inspection) \\
\midrule
\textbf{Reasoning Inspector}
  & \texttt{<error\_stage>},
  & \textbf{Solver} (retry) / \\
  & \texttt{<explanation>}
  & Answer Selection \\
\midrule
\textbf{Memoizer}
  & hit/miss + cached answer
  & Pipeline control \\
  & (similarity $\geq \tau_m$)
  & (fast path) \\
\bottomrule
\end{tabular*}
\end{table}

\begin{algorithm}[t]
\caption{PRISMA: Plan-Retrieve-Inspect-Solve-Memoize}
\label{alg:prisma}
\begin{algorithmic}[1]
\scriptsize
\setlength{\baselineskip}{0.92\baselineskip}
\Require Question set $\mathcal{Q}$, memory threshold $\tau_m$, max expansions $E_r$ (ctx) and $E_p$ (post hoc), doc limits $k_{\max}$ and $k_{\text{solve}}$
\Ensure Predictions $\mathcal{Y}$, full traces $\mathcal{T}$, inspector datasets $(\mathcal{C}_{\text{ctx}}, \mathcal{C}_{\text{rea}})$

\State \textbf{Initialize:} empty run-scoped memory cache $\mathcal{M}$
\State $\mathcal{Y},\mathcal{T},\mathcal{C}_{\text{ctx}},\mathcal{C}_{\text{rea}}\gets[\ ]$

\For{each question $q \in \mathcal{Q}$}
  \State $\mathcal{T}_q\gets\text{NewTrace}(q)$
  \Statex \textit{Each Inspector call appends its audit to the corresponding $\mathcal{C}$ and to $\mathcal{T}_q$.}
  \Statex \textbf{// Phase 1: Plan}
  \State $\mathcal{S} \gets \text{Planner}(q)$; $\mathcal{A} \gets [\ ]$

  \For{each subquestion $s \in \mathcal{S}$}
    \Statex \textbf{// Cascade Guard + Fill Placeholders}
    \If{$\text{UsesRejectedAnswer}(s,\mathcal{A})$}
      \State $s \gets \text{RewriteSubquestion}(q,s,\mathcal{A})$
      \If{$\text{UsesRejectedAnswer}(s,\mathcal{A})$} \State $\mathcal{A}.\text{append}(\text{``Not found in the documents''})$; \textbf{break} \EndIf
    \EndIf
    \State $s \gets \text{FillPlaceholders}(s,\mathcal{A})$

    \Statex \textbf{// Step 1: Memoize + Retrieve}
    \State $\text{hit} \gets \text{MemoryLookup}(\mathcal{M}, s, \tau_m)$
    \If{$\text{hit} \neq \emptyset$}
      \State $a \gets \text{hit.answer}$; $\mathcal{A}.\text{append}(a)$ \Comment{HIT: output answer directly}
      \State \textbf{continue}
    \EndIf

    \State $\mathcal{E}_0 \gets \text{Retrieve}(s)$ \Comment{Three-stage retrieval cascade}

    \Statex \textbf{// Step 2: Context Inspector + Repair}
    \State $r_{\text{ctx}} \gets \text{Context Inspector}(q,s,\mathcal{E}_0,\mathcal{A})$ \Comment{Save ctx sample \textit{for every call}}
    \If{$r_{\text{ctx}}.\text{error\_stage} = \texttt{subquestion}$}
      \State $s \gets \text{RewriteSubquestion}(q,s,r_{\text{ctx}}.\text{explanation},\mathcal{A})$; $\mathcal{E}_0 \gets \text{Retrieve}(s)$
      \State $r_{\text{ctx}} \gets \text{Context Inspector}(q,s,\mathcal{E}_0,\mathcal{A})$
    \EndIf
    \For{$t=1$ to $E_r$}
      \If{$r_{\text{ctx}}.\text{error\_stage} \neq \texttt{retrieval}$} \State \textbf{break} \EndIf
      \State $q' \gets \text{RewriteQuery}(q,s,r_{\text{ctx}}.\text{explanation})$
      \State $\mathcal{E}_0 \gets \text{MergeUnique}(\text{Retrieve}(q'), \mathcal{E}_0)[:k_{\max}]$; $r_{\text{ctx}} \gets \text{Context Inspector}(q,s,\mathcal{E}_0,\mathcal{A})$
    \EndFor
    \State $\mathcal{E} \gets \text{SelectDocs}(\mathcal{E}_0, k_{\text{solve}})$

    \Statex \textbf{// Step 3: Solve}
    \State $\text{exec} \gets \text{Solver}(s,\mathcal{E})$; $a \gets \text{ExtractAnswer}(\text{exec})$

    \Statex \textbf{// Step 4: Reasoning Inspector + Repair}
    \State $r_{\text{rea}} \gets \text{Reasoning Inspector}(q,s,\mathcal{E},\text{exec},a)$ \Comment{Save rea sample \textit{for every call}}
    \If{$r_{\text{rea}}.\text{error\_stage} = \texttt{none}$}
      \State $r_{\text{rea}} \gets \text{ExecutionGuardIfNeeded}(s,a,\mathcal{E})$
    \EndIf
    \For{$t=1$ to $E_p$}
      \If{$\neg(r_{\text{rea}}.\text{error\_stage} = \texttt{reasoning} \land \text{LooksMissingEvidence}(r_{\text{rea}}))$} \State \textbf{break} \EndIf
      \State $q' \gets \text{RewriteQuery}(q,s,r_{\text{rea}}.\text{explanation})$
      \State $\mathcal{E} \gets \text{MergeUnique}(\text{Retrieve}(q'),\mathcal{E})[:k_{\max}]$; $\text{exec} \gets \text{Solver}(s,\mathcal{E})$; $a \gets \text{ExtractAnswer}(\text{exec})$
      \State $r_{\text{rea}} \gets \text{Reasoning Inspector}(q,s,\mathcal{E},\text{exec},a)$
    \EndFor
    \If{$r_{\text{rea}}.\text{error\_stage} \in \{\texttt{reasoning}, \texttt{extraction}\}$}
      \State $\text{exec}' \gets \text{SolverWithFeedback}(s,\mathcal{E},r_{\text{rea}},\text{exec})$; $a' \gets \text{ExtractAnswer}(\text{exec}')$
      \State $a \gets \text{CompareAnswers}(a,a',s,\mathcal{E},r_{\text{rea}})$ \Comment{Conservative selection}
    \EndIf

    \Statex \textbf{// Step 5: Memoize Store}
    \State $\text{MemoryStore}(\mathcal{M}, s, a)$; $\mathcal{A}.\text{append}(a)$ \Comment{Record step info into $\mathcal{T}_q$}
  \EndFor

  \State $a_q \gets \mathcal{A}[-1]$; $\mathcal{Y}.\text{append}(a_q)$; $\mathcal{T}.\text{append}(\mathcal{T}_q)$
  \State $\text{SaveTrace}(\mathcal{T}_q)$ \Comment{Persist full per-sample trace}
\EndFor

\State $\text{Clear}(\mathcal{M})$ \Comment{Discard cache after the complete evaluation run}
\State \Return $\mathcal{Y},\mathcal{T},\mathcal{C}_{\text{ctx}},\mathcal{C}_{\text{rea}}$
\end{algorithmic}
\end{algorithm}

\subsection{Architecture Overview}
\label{sec:architecture_overview}

As illustrated in Fig.~\ref{fig:motivation}, PRISMA processes each query through two phases.

\noindent\textbf{Phase 1: Planning.} The Planner decomposes the query $q$ into dependency-aware subquestions $[SQ_1, SQ_2, \ldots, SQ_n]$, where later subquestions may reference earlier answers via placeholders. For example, given the multi-hop query ``What is the capital of the country that Lyon belongs to?'', the Planner may first generate ``Which country does the city of Lyon belong to?'' and then ask ``What is the capital of [ANSWER\_0]?'' so that the second subquestion is explicitly conditioned on the intermediate answer from the first step.

\noindent\textbf{Phase 2: Iterative Solving with Inspection.} For each subquestion $SQ_i$, the pipeline executes a Retrieve-Inspect-Solve-Inspect loop:
\begin{enumerate}[leftmargin=*]
\item The \emph{Memoizer} checks its semantic cache; on cache miss, the \emph{Retriever} fetches passages via a three-stage cascade (dense $\rightarrow$ hybrid $\rightarrow$ cross-encoder).
\item The \emph{Context Inspector} validates subquestion quality and document sufficiency. Subquestion issues trigger Planner rewrites; retrieval failures trigger cumulative expansion that aggregates unique documents across rounds.
\item The \emph{Solver} produces a grounded answer with explicit source citations following its output schema.
\item The \emph{Reasoning Inspector} verifies answer grounding against cited evidence. Detected issues may invoke post hoc retrieval expansion or answer retry; the system then compares original vs.\ retry answers using Inspector signals and selects conservatively.
\end{enumerate}

\noindent Before executing each $SQ_i$, the Memoizer runs cascade guards to detect dependencies on rejected or missing placeholders (e.g., ``[ANSWER\_1] = Not found in the documents''); if triggered, it rewrites the subquestion or blocks execution to prevent error propagation. Algorithm~\ref{alg:prisma} provides the complete pseudocode.

\begin{figure*}[!t]
\centering
\includegraphics[width=\textwidth]{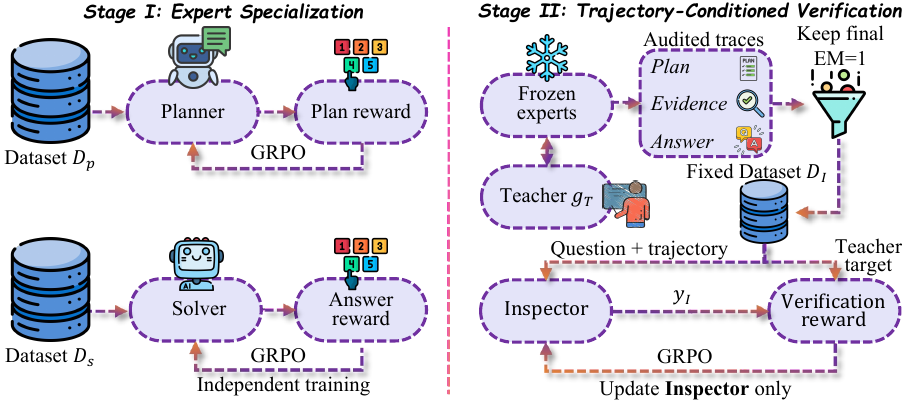}
\caption{Decoupled two-stage training of \textsc{PRISMA}. Stage~I independently specializes the Planner on $\mathcal{D}_P$ with plan rewards and the Solver on $\mathcal{D}_S$ with answer rewards, using GRPO in both branches. The resulting experts are then frozen. Stage~II runs the frozen experts with a high-capacity teacher Inspector $g_T$ to produce audited trajectories containing plans, evidence, and answers; only rollouts with correct final answers (EM$=1$) are retained to form the fixed Inspector dataset $\mathcal{D}_I$. The Inspector is trained on these question--trajectory pairs with teacher audit targets, so verification is learned over a stable expert pipeline while the Planner and Solver remain fixed.}
\label{fig:two_stage_training}
\end{figure*}
\subsection{Agent Roles and Functional Design}
\label{sec:agent_architecture}

\subsubsection{Planner}
The \textbf{Planner} decomposes a multi-hop query $q$ into a dependency-aware chain of subquestions $[SQ_1, \ldots, SQ_n]$ following the XML plan schema (Table~\ref{tab:agent_schema}). Later subquestions may reference earlier answers via typed placeholders (e.g., \texttt{[ANSWER\_1]}), encoding the dependency structure of the evidence chain. The Planner's output quality directly determines retrieval precision: a well-decomposed subquestion isolates a single bridge entity, enabling the Retriever to locate its supporting passage from a large-scale open-domain corpus.

\subsubsection{Inspector (Dual-Phase Verification)}
The \textbf{Inspector} operates in two phases, both producing structured audit outputs following the same XML schema but targeting different failure modes.

\noindent\textbf{Context Inspector} validates (i) subquestion quality (resolution, grammar, entity binding) and (ii) document sufficiency before solving. When \texttt{error\_stage = subquestion}, it triggers the Planner to rewrite the subquestion. When \texttt{error\_stage = retrieval}, it triggers cumulative retrieval expansion that generates alternative queries from the Inspector's explanation and aggregates unique documents across rounds up to a preset document budget.

\noindent\textbf{Reasoning Inspector} validates (i) answer grounding against cited evidence, (ii) entity resolution correctness, and (iii) answer-type alignment (e.g., who$\rightarrow$person, when$\rightarrow$date). Detected issues may invoke post hoc retrieval expansion or answer retry with feedback. The system then compares original vs.\ retry answers using Inspector signals and execution guard heuristics, selecting conservatively.

\subsubsection{Solver}
The \textbf{Solver} generates an answer for each subquestion from the retrieved evidence, producing structured output with explicit reasoning traces, source citations, and an extracted answer (Table~\ref{tab:agent_schema}). The Solver selects evidence conditionally: if retries are not exhausted, it uses the newest documents from expansion; otherwise, it falls back to the earliest documents. The citation requirement in the output schema forces the Solver to ground every claim in specific passages, enabling the downstream Reasoning Inspector to verify faithfulness.

\subsection{Retriever and Memoizer}
\label{sec:tools}

\subsubsection{Three-Stage Retrieval Cascade}
The \textbf{Retriever} targets evidence coverage via a three-stage cascade. It first recalls a broad candidate pool with dense retrieval, then applies hybrid dense-sparse filtering:
\begin{equation}
s_{\text{hyb}}(sq,d) = \alpha\, s_{\text{dense}}(sq,d) + (1-\alpha)\, s_{\text{sparse}}(sq,d),
\end{equation}
where $sq$ is the subquestion, $d$ is a passage, $s_{\text{dense}}$ and $s_{\text{sparse}}$ are normalized dense and sparse relevance scores, and $\alpha \in [0,1]$ balances them.

\noindent\textbf{Stage 1 (Dense):} A dense retriever maps the subquestion and corpus passages into a shared embedding space and retrieves semantically relevant candidates.

\noindent\textbf{Stage 2 (Hybrid):} Dense semantic relevance is combined with sparse lexical evidence to retain candidates that are both semantically aligned and lexically grounded.

\noindent\textbf{Stage 3 (Reranking):} A reranker scores query-document pairs through direct interaction and selects the evidence passed to the Solver.

\noindent On insufficient evidence (as determined by the Context Inspector), the system generates alternative queries from Inspector feedback and fetches additional documents via the same cascade. New documents merge with previous batches using ID-based deduplication until a preset expansion budget or document budget is reached. Concrete retrieval backbones and hyperparameters are reported in Section~\ref{sec:experimental_setup}.


\subsubsection{Memoizer}
The \textbf{Memoizer} targets efficiency by reusing semantically matched answers and recording execution traces. It consists of two components:

\noindent\textbf{Evidence Store:} Run-scoped in-memory cache of subquestion--answer pairs. The cache is initialized empty at the beginning of an evaluation run, shared across all questions in that run, and populated online as subquestions are solved. Consequently, a later question can directly reuse an answer when its subquestion semantically matches one solved earlier in the same run. Cache hits are determined via encoder-based semantic similarity:
\begin{equation}
\text{hit}(sq)=\mathbb{I}\left[\max_{(sq_i,a_i)\in \mathcal{M}} \cos(e(sq),e(sq_i)) \ge \tau_m \right],
\end{equation}
where $\mathcal{M}$ contains subquestions and answers accumulated earlier in the current run, $e(\cdot)$ is the cache encoder, $\tau_m$ is the similarity threshold, and the matched $a_i$ is returned when $\text{hit}(sq)=1$. We define the maximum over an empty cache as $-\infty$, so the first lookup is necessarily a miss. The cache is discarded after the complete run, so no entries are carried into an independent evaluation while repeated or semantically equivalent subquestions within the same run benefit from an increasingly warm cache.

\noindent\textbf{Trajectory Store:} Persistent logging of execution events with metadata including event type (plan, retrieve, inspect, solve, cache\_hit, retry, error), timestamps for latency analysis, complete input/output payloads, and parent/child relationships for trace reconstruction.

\subsection{Decoupled Training for Multi-Agent Coordination}
\label{sec:training}

Fig.~\ref{fig:two_stage_training} summarizes the specialize--freeze--verify procedure. Stage~I independently trains the Planner and Solver with agent-specific rewards. Stage~II freezes both experts, constructs a fixed dataset of teacher-audited trajectories, and updates only the Inspector to learn trajectory-conditioned verification and recovery. Both stages use GRPO (Section~\ref{sec:grpo_preliminary}).

\subsubsection{Stage I: Expert Specialization}
\label{sec:stage1}

\paragraph{Notation}
$x$ denotes an input question; $y_P$ a Planner plan (a structured sequence of subquestions);
$E$ the retrieved evidence; $y_S$ a Solver output (a structured answer with citations);
$y_I$ an Inspector audit output; $a^*$ the gold answer; $\mathcal{S}^*$ the gold supporting-document set; and $e^*$ the gold audit label.
$\mathcal{D}_P$ and $\mathcal{D}_S$ denote the Planner and Solver training distributions, respectively.
$\theta_P$, $\theta_S$, and $\theta_I$ denote the learnable parameters of the Planner, Solver, and Inspector, respectively, with $\pi_{\theta}(\cdot \mid \cdot)$ the corresponding policy (as defined in Section~\ref{sec:grpo_preliminary}).

\paragraph{Objectives}
We optimize the Planner and Solver \emph{independently} to induce functional specialization (decomposition vs.\ grounded answering):
\begin{equation}
\theta_P^* = \arg\max_{\theta_P}\
\mathbb{E}_{x\sim\mathcal{D}_P}\
\mathbb{E}_{y_P \sim \pi_{\theta_P}(\cdot \mid x)}
\!\left[ R_{\text{plan}}(x,y_P) \right],
\label{eq:stage1_planner_obj}
\end{equation}

\begin{equation}
\begin{aligned}
\theta_S^*
&= \arg\max_{\theta_S}\;
\mathbb{E}_{(x,E,a^*,\mathcal{S}^*)\sim\mathcal{D}_S}\\
&\quad
\mathbb{E}_{y_S \sim \pi_{\theta_S}(\cdot \mid x, E)}\\
&\quad
\left[ R_{\text{solve}}(y_S; a^*, \mathcal{S}^*) \right].
\end{aligned}
\label{eq:stage1_solver_obj}
\end{equation}
\noindent After Stage~I, we denote the calibrated expert policies as:
\begin{equation}
\text{System}^{(1)} := (\pi_{\theta_P^*},\,\pi_{\theta_S^*}).
\label{eq:system1_def}
\end{equation}

\subsubsection{Stage II: Trajectory-Conditioned Verification}
\label{sec:stage2}

Stage~II updates only the Inspector parameters $\theta_I$ on a fixed, offline-constructed trajectory dataset, keeping both policies in $\text{System}^{(1)}$ frozen. The Inspector learns to audit residual errors from question--trajectory observations, with teacher audit targets used in the verification reward. This decoupling is a deliberate variance-control mechanism: under joint training, updates to $\theta_P$ and $\theta_S$ shift the trajectory distribution seen by the Inspector and introduce an additional source of gradient variability.

\begin{proposition}[Elimination of Upstream-Induced Gradient Variability]
\label{prop:variance}
Let $\Theta=(\theta_P,\theta_S)$ denote the upstream policy state and let $G(\tau,\Theta)$ be a stochastic estimator of the Inspector gradient obtained from trajectory $\tau$. For a vector-valued gradient, define its total variance as $\mathcal{V}(G):=\mathbb{E}[\lVert G-\mathbb{E}[G]\rVert_2^2]$. If joint training exposes the Inspector to a distribution of evolving upstream states $\Theta$, the law of total variance gives
\begin{equation}
\begin{aligned}
\mathcal{V}(G_{\mathrm{joint}})
= {} & \underbrace{\mathbb{E}_{\Theta}\!\left[\mathcal{V}(G\mid\Theta)\right]}_{\text{within-state stochasticity}} \\
& + \underbrace{\mathcal{V}_{\Theta}\!\left(\mathbb{E}[G\mid\Theta]\right)}_{\text{upstream-induced variability}}.
\end{aligned}
\label{eq:variance_decomp}
\end{equation}
The second term is nonnegative and captures variation in the expected Inspector gradient caused exclusively by changes in the upstream policies. Stage~II fixes $\Theta=\Theta^*:=(\theta_P^*,\theta_S^*)$, making this entire between-state component identically zero:
\begin{equation}
\mathcal{V}_{\Theta}\!\left(\mathbb{E}[G\mid\Theta^*]\right)=0,\qquad
\mathcal{V}(G_{\mathrm{frozen}})=\mathcal{V}(G\mid\Theta^*).
\label{eq:variance_frozen}
\end{equation}
Thus, decoupled training removes the full upstream-induced component of gradient variability, leaving only stochasticity under a fixed expert pipeline.
\end{proposition}

\noindent With a stationary observation distribution, the Inspector can learn stable, stage-specific error patterns rather than continually adapting to moving targets. Concretely, running $\text{System}^{(1)}$ on $x$ produces an expert trajectory:
\begin{equation}
\tau_{P,S} = \{y_P, E, y_S\},
\label{eq:traj_def}
\end{equation}
The Inspector conditions on this trace via an augmented observation:
\begin{equation}
s_{\mathrm{aug}} = (x, \tau_{P,S}),
\label{eq:aug_state}
\end{equation}
and is optimized with GRPO over the fixed success-conditioned trajectory dataset $\mathcal{D}_I(\text{System}^{(1)})$ defined below. For compactness, write this dataset as $\mathcal{D}_I$ in the objective:
\begin{equation}
\begin{aligned}
\theta_I^*
&= \arg\max_{\theta_I} \mathcal{J}_I(\theta_I),\\
\mathcal{J}_I(\theta_I)
&= \mathbb{E}_{(s_{\mathrm{aug}},e^*)\sim\mathcal{D}_I}
\mathbb{E}_{y_I \sim \pi_{\theta_I}(\cdot \mid s_{\mathrm{aug}})}\\
&\quad \left[ R_{\text{verify}}(y_I; e^*) \right].
\end{aligned}
\label{eq:stage2_inspector_obj}
\end{equation}
This yields the complete system:
\begin{equation}
\begin{aligned}
\text{System}^{(2)}
&:= \text{System}^{(1)} \oplus \pi_{\theta_I^*}
= (\pi_{\theta_P^*},\,\pi_{\theta_S^*},\,\pi_{\theta_I^*}),
\end{aligned}
\label{eq:system2_def}
\end{equation}
where $\oplus$ denotes augmenting the calibrated experts with a verification policy that observes $s_{\mathrm{aug}}$ and triggers recovery when needed.

\paragraph{Success-conditioned trajectory distillation and teacher audit labels}
Stage~II supervision is defined at the \emph{audit level} rather than by end-to-end EM, which is affected by confounded factors beyond audit quality (corpus coverage, retriever recall, Solver robustness). We run $\text{System}^{(1)}$ with a high-capacity oracle Inspector teacher $g_T$ and select successful traces:
\begin{equation}
\mathcal{X}^{+}=\{x\in\mathcal{X}\;:\;\mathbb{I}[\phi(\hat{a}^{(T)}(x))=\phi(a^*(x))]=1\},
\end{equation}
where $\hat{a}^{(T)}(x)$ is the final answer produced under $g_T$ and $\phi(\cdot)$ is answer normalization. For each $x \in \mathcal{X}^{+}$, the teacher generates audit targets $e^*(x) = g_T(s_{\mathrm{aug}}(x))$, yielding the Stage~II training dataset:
\begin{equation}
\mathcal{D}_I(\text{System}^{(1)}) = \{(s_{\mathrm{aug}}(x), e^*(x)) : x \in \mathcal{X}^{+}\}.
\end{equation}
This success-conditioned construction retains trajectories whose teacher-guided intervention demonstrates verified end-to-end utility, filtering unsuccessful or unrecoverable teacher rollouts before optimization. It therefore turns verification training into a trajectory-distillation problem with a fixed observation distribution and high-quality, expert-level audit supervision.

\subsubsection{Recovery-Gain Analysis}
\label{sec:gain_decomposition}

Let $S^{(k)}\in\{0,1\}$ denote the end-to-end success indicator of $\text{System}^{(k)}$ on question $x$.
The Stage-II gain decomposes as:
\begin{equation}
\begin{aligned}
\mathbb{E}[S^{(2)}] = {} & \mathbb{E}[S^{(1)}] \\
& + \Pr(S^{(1)}\!=\!0)\Pr(\texttt{rec}\!=\!1 \mid S^{(1)}\!=\!0) \\
& - \Pr(S^{(1)}\!=\!1)\Pr(\texttt{reg}\!=\!1 \mid S^{(1)}\!=\!1),
\end{aligned}
\label{eq:two_stage_gain}
\end{equation}
where $\texttt{rec}=1$ (recovered) indicates converting a failing trajectory to a correct answer via Inspector-triggered recovery, and $\texttt{reg}=1$ (regressed) represents a do-no-harm violation where an initially correct trajectory is degraded by unnecessary intervention. This decomposition yields a do-no-harm-aware criterion: Stage~II improves end-to-end performance if and only if $\Pr(S^{(1)}\!=\!0)\Pr(\texttt{rec}\!=\!1\mid S^{(1)}\!=\!0) > \Pr(S^{(1)}\!=\!1)\Pr(\texttt{reg}\!=\!1\mid S^{(1)}\!=\!1)$, or equivalently, if recovered failures outnumber regressed successes. Our empirical 13-to-2 recovery--regression outcome satisfies this criterion by a factor of 6.5 (Table~\ref{tab:inspection_impact}).

\subsubsection{Agent-Specific Reward Design}
\label{sec:reward_design}

Each agent's reward is designed within its schema-defined responsibility: Stage~I rewards calibrate \emph{capabilities} (decomposition quality for Planner, grounded answering for Solver), while Stage~II rewards calibrate \emph{verification reliability} conditioned on expert trajectories. We use $w_{\cdot}^{P}$, $w_{\cdot}^{S}$, and $w_{\cdot}^{I}$ to denote positive scalar weights for each agent. Appendix~\ref{sec:training_config} provides the training configuration and reward weights. Appendix~\ref{sec:implementation} lists the agent prompts.

\paragraph{Planner Reward}
The Planner is optimized to generate dependency-aware subquestions that preserve query constraints and boost retrievability:
\begin{equation}
\begin{aligned}
R_{\text{plan}}(x,y_P)
&= w_{\mathrm{fmt}}^{P}\,r_{\text{fmt}}^{P}(y_P)
 + w_{\mathrm{cnt}}^{P}\,r_{\text{count}}(x,y_P) \\
&\quad + w_{\mathrm{judge}}^{P}\sum_{i=1}^{4} r_{\text{judge}}^{(i)}(x,y_P),
\end{aligned}
\label{eq:r_plan_main}
\end{equation}
where $r_{\text{fmt}}^{P}$ measures output format validity, $r_{\text{count}}$ encourages appropriate hop granularity, and $r_{\text{judge}}^{(i)}$ denotes semantic planning criteria such as qualifier preservation, entity binding, disambiguation, and dependency logic.

\paragraph{Solver Reward}
The Solver is optimized to answer from retrieved evidence with faithful citations:
\begin{equation}
\begin{aligned}
R_{\text{solve}}(y_S; a^*, \mathcal{S}^*)
&= w_{\mathrm{fmt}}^{S}\,r_{\text{fmt}}^{S}(y_S)
 + w_{\mathrm{acc}}^{S}\,r_{\text{acc}}(y_S,a^*) \\
&\quad + w_{\mathrm{rel}}^{S}\,r_{\text{rel}}(y_S,\mathcal{S}^*),
\end{aligned}
\label{eq:r_solve_main}
\end{equation}
where $r_{\text{fmt}}^{S}$ measures output format validity, $r_{\text{acc}}$ rewards normalized exact match against the gold answer, and $r_{\text{rel}}$ measures citation faithfulness against the gold supporting-document set.

\paragraph{Inspector Reward}
The Inspector is optimized for verification accuracy and targeted recovery:
\begin{equation}
\begin{aligned}
R_{\text{verify}}(y_I; e^*)
&= w_{\mathrm{fmt}}^{I}\,r_{\text{fmt}}^{I}(y_I)
 + w_{\mathrm{det}}^{I}\,r_{\text{detect}}(y_I,e^*) \\
&\quad + w_{\mathrm{len}}^{I}\,r_{\text{length}}(y_I,e^*),
\end{aligned}
\label{eq:r_verify_main}
\end{equation}
where $r_{\text{fmt}}^{I}$ measures output format validity, $r_{\text{detect}}$ rewards error-stage correctness, and $r_{\text{length}}$ shapes the explanation toward concise no-error outputs and sufficiently informative error diagnoses.


\begin{table}[!t]
\caption{Representative Examples from the Human-Curated Domain-Specific Sets}
\label{tab:domain_examples}
\centering
\scriptsize
\setlength{\tabcolsep}{2pt}
\begin{tabularx}{\columnwidth}{@{}l l X l@{}}
\toprule
\textbf{Domain} & \textbf{Form} & \textbf{Example Question} & \textbf{Answer} \\
\midrule
Chemistry & Bridge & The element with atomic number 79 was historically sought by which group of medieval scientists? & Alchemists \\
Food & Bridge & The inventor of Coca-Cola was a pharmacist in which US state? & Georgia \\
Game & Bridge & The company that created Super Mario Bros is headquartered in which Japanese city? & Kyoto \\
Geography & Bridge & The longest river in Africa flows through which country's capital city? & Egypt \\
Music & Comparison & Is The Beatles or The Rolling Stones the older band? & The Beatles \\
\bottomrule
\end{tabularx}
\end{table}

\begin{table}[!t]
\caption{Evaluation Benchmarks (3 In-Distribution, 7 Out-of-Distribution)}
\label{tab:dataset_full_stats}
\centering
\scriptsize
\begin{tabular*}{\columnwidth}{@{\extracolsep{\fill}}lll@{}}
\toprule
\textbf{Dataset} & \textbf{Domain} & \textbf{Type} \\
\midrule
\multicolumn{3}{c}{\textbf{\textit{In-Distribution Benchmarks}}} \\
\midrule
HotpotQA   & Wikipedia  & Comparison, Bridge \\
2WikiMHQA  & Wikipedia  & Compositional \\
MuSiQue    & Wikipedia  & Adversarial \\
\midrule
\multicolumn{3}{c}{\textbf{\textit{Out-of-Distribution Benchmarks}}} \\
\midrule
NQ         & Wikipedia   & Factoid \\
Bamboogle  & Wikipedia   & Adversarial \\
Chemistry  & Scientific  & Technical \\
Food       & Culinary    & Commonsense \\
Game       & Entertainment & Pop culture \\
Geography  & Spatial     & Location \\
Music      & Arts        & Creative \\
\bottomrule
\end{tabular*}
\end{table}

\section{Experiment}

\subsection{Experimental Setup}
\label{sec:experimental_setup}

\paragraph{Datasets}
We evaluate \textsc{PRISMA} on \textbf{ten open-domain} QA benchmarks (Table~\ref{tab:dataset_full_stats}). This includes three \textbf{in-distribution} benchmarks---HotpotQA~\cite{yang2018hotpotqa} (90K), 2WikiMultiHopQA~\cite{ho2020constructing} (150K), and MuSiQue~\cite{trivedi2022musique} (25K)---and seven \textbf{out-of-distribution} benchmarks: Natural Questions (NQ)~\cite{kwiatkowski2019natural} (8K), Bamboogle~\cite{press2023measuring} (125), and five human-curated domain-specific sets (Chemistry/Food/Game/Geography/Music; 150 each). For datasets with more than 500 questions, we sample 500 from Dev. The five domain-specific datasets are manually authored by the research team, each containing 150 multi-hop questions requiring bridge entity chaining or comparison reasoning.

\noindent\textbf{Domain-Specific Set Construction and Quality Control.} To assess out-of-distribution transfer beyond the topic and template biases of the in-distribution benchmarks, we construct five domain-specific evaluation sets covering Chemistry, Food, Game, Geography, and Music, with 150 questions per domain. Questions are authored around factual relations represented in English Wikipedia and require either bridge-entity chaining or comparison reasoning. Each record includes a gold answer and accepted aliases for EM/F1 evaluation, but provides neither gold passages nor gold decompositions, requiring every method to retrieve and compose evidence from the same 21M-passage corpus. We apply deterministic integrity checks for required fields, answerability, reasoning-type validity, and duplicate IDs or normalized questions. Two domain experts then perform a final manual review for question clarity, answer correctness, and validity of the intended multi-hop relation. All 750 records pass these checks. Representative examples are shown in Table~\ref{tab:domain_examples}.

\begin{table}[!t]
\caption{Evidence Provision and Retrieval Settings. Local Retrieval Experiments Use the DPR 21M-Passage Corpus~\cite{karpukhin2020dense}. $^\dagger$Original-Paper Results Retain Their Source Settings}
\label{tab:eval_setting_comparison}
\centering
\scriptsize
\setlength{\tabcolsep}{2pt}
\begin{tabular*}{\columnwidth}{@{\extracolsep{\fill}}p{2.2cm}p{3.2cm}p{1.8cm}@{}}
\toprule
\textbf{Method} & \textbf{Description} & \textbf{Setting Detail} \\
\midrule
\multicolumn{3}{c}{\textbf{\textit{Reader Setting (Not Included in Main Results)}}} \\
\midrule
ARENA~\cite{ren2025arena} & Reader over provided evidence & 10--20 paragraphs \\
\midrule
\multicolumn{3}{c}{\textbf{\textit{Corpus Retrieval (Not Included in Main Results)}}} \\
\midrule
MMOA~\cite{chen2025mmoa} & Wikipedia retrieval before selection & $K$=10 retrieved docs \\
OPERA~\cite{liu2025opera} & Planning-aware RAG & Corpus retrieval \\
RT-RAG~\cite{shi2026reasoning} & Tree-structured retrieval and reasoning & Retrieved context: 3000 tokens \\
\midrule
\multicolumn{3}{c}{\textbf{\textit{Retrieval Systems in the Main Comparison}}} \\
\midrule
IRCoT~\cite{trivedi2023interleaving} & Interleaved CoT retrieval & 21M \\
RAG-Star~\cite{jiang2025rag} & Deliberative reasoning with retrieval verification & 21M \\
Search-o1~\cite{li2025search} & Agentic search in reasoning & 21M \\
Retrobust~\cite{yoran2024robust} & SFT for retrieval robustness & 21M \\
RAG-DDR~\cite{li2025ragddr} & RL-based retrieval-reasoning optimization & 21M \\
QAgent$^\dagger$~\cite{jiang2025qagent} & RL-trained multi-step agent & Source protocol \\
TIRESRAG-R1$^\dagger$~\cite{he2025tiresrag} & Think-retrieve-reflect with RL & Source protocol \\
\textbf{PRISMA} & \textbf{Schema-coordinated multi-agent with in-loop verification} & \textbf{21M} \\
\bottomrule
\end{tabular*}
\end{table}

\begin{table}[!t]
\caption{Retrieval and Cache Implementation Details}
\label{tab:retrieval_config}
\centering
\small
\begin{tabularx}{\columnwidth}{lX}
\toprule
\textbf{Parameter} & \textbf{Configuration} \\
\midrule
Corpus & DPR Wikipedia (21M passages) \\
Dense retriever & BGE-M3 with FAISS IVF (nlist=4096, nprobe=128) \\
Stage-1 candidates & $k_1 = 100$ documents \\
Hybrid fusion & $\alpha = 0.65$ (65\% dense, 35\% sparse) \\
Stage-2 candidates & $k_2 = 30$ documents \\
Reranker & BGE-reranker-v2-m3 \\
Final evidence & top-10 documents \\
Expansion budget & 3 iterations; 20 documents per subquestion \\
Memoize threshold & BGE-M3 cosine similarity, $\tau_m=0.85$ \\
\bottomrule
\end{tabularx}
\end{table}

\begin{table*}[!t]
\caption{Main Results (EM/F1). \uwave{Wavy}: Best Closed-Source API. \underline{Underline}: Best Open-Source (Excl.\ PRISMA). $^\dagger$Original Paper Results. $^\ast$Best of IRCoT, RAG-Star, Search-o1 Per Backbone}
\label{tab:main_results}
\centering
\scriptsize
\setlength{\tabcolsep}{2pt}

\begin{tabular*}{\textwidth}{@{\extracolsep{\fill}}lccccccc@{}}
\toprule
\textit{In-Distribution Evaluation} & & \multicolumn{2}{c}{\textbf{MuSiQue}~\cite{trivedi2022musique}} & \multicolumn{2}{c}{\textbf{HotpotQA}~\cite{yang2018hotpotqa}} & \multicolumn{2}{c}{\textbf{2WikiMHQA}~\cite{ho2020constructing}} \\
\textbf{Method} & \textbf{Type} & EM & F1 & EM & F1 & EM & F1 \\
\midrule
\multicolumn{8}{c}{\textbf{\textit{Closed-book (No Retrieval)}}} \\
\midrule
\quad Qwen3-4B~\cite{qwen3} & Pre-Trained & 2.0 & 5.6 & 13.4 & 19.8 & 28.6 & 32.9 \\
\quad Qwen3-8B~\cite{qwen3} & Pre-Trained & 6.4 & 12.5 & 17.4 & 25.5 & 8.2 & 15.0 \\
\quad Qwen3-30B~\cite{qwen3} & Pre-Trained & 2.6 & 11.8 & 20.8 & 30.4 & 29.4 & 35.5 \\
\quad DeepSeek-V3.2~\cite{liu2025deepseek} & Pre-Trained (API) & 11.8 & 25.5 & 39.2 & 50.3 & 16.6 & 26.7 \\
\quad Gemini-2.5-Flash~\cite{google2025gemini25flash} & Pre-Trained (API) & 15.8 & 27.0 & 38.4 & 51.4 & 28.0 & 41.3 \\
\quad GPT-5-Medium~\cite{openai2025gpt5} & Pre-Trained (API) & 9.8 & 20.7 & 27.4 & 38.5 & 6.4 & 20.5 \\
\midrule
\multicolumn{8}{c}{\textbf{\textit{Single-Step RAG}}} \\
\midrule
\quad Qwen3-4B~\cite{qwen3} & Pre-Trained & 8.4 & 14.9 & 29.4 & 37.5 & 13.8 & 18.5 \\
\quad Qwen3-8B~\cite{qwen3} & Pre-Trained & 8.6 & 12.8 & 29.0 & 34.9 & 6.0 & 9.8 \\
\quad Qwen3-30B~\cite{qwen3} & Pre-Trained & 9.4 & 19.6 & 35.2 & 45.7 & 31.0 & 38.1 \\
\quad DeepSeek-V3.2~\cite{liu2025deepseek} & Pre-Trained (API) & 10.4 & 16.5 & 20.6 & 30.7 & 25.2 & 41.1 \\
\quad Gemini-2.5-Flash~\cite{google2025gemini25flash} & Pre-Trained (API) & 11.0 & 18.3 & 19.2 & 29.9 & 16.4 & 26.6 \\
\quad GPT-5-Medium~\cite{openai2025gpt5} & Pre-Trained (API) & 15.2 & 27.6 & 36.6 & 50.5 & 17.0 & 31.3 \\
\midrule
\multicolumn{8}{c}{\textbf{\textit{Multi-Step RAG$^\ast$ (IRCoT~\cite{trivedi2023interleaving}, RAG-Star~\cite{jiang2025rag}, Search-o1~\cite{li2025search})}}} \\
\midrule
\quad Qwen3-4B~\cite{qwen3} & Pre-Trained & 7.6 & 14.6 & 27.2 & 34.7 & 34.8 & 38.6 \\
\quad Qwen3-8B~\cite{qwen3} & Pre-Trained & 6.8 & 11.8 & 24.2 & 30.0 & 15.0 & 18.9 \\
\quad Qwen3-30B~\cite{qwen3} & Pre-Trained & 14.2 & 22.3 & 35.8 & 46.0 & 36.0 & 43.3 \\
\quad DeepSeek-V3.2~\cite{liu2025deepseek} & Pre-Trained (API) & 18.6 & 31.8 & \uwave{54.8} & \uwave{66.0} & 57.8 & 66.2 \\
\quad Gemini-2.5-Flash~\cite{google2025gemini25flash} & Pre-Trained (API) & 22.6 & 33.7 & 54.6 & 65.2 & \uwave{59.6} & \uwave{70.6} \\
\quad GPT-5-Medium~\cite{openai2025gpt5} & Pre-Trained (API) & \uwave{23.4} & \uwave{35.0} & 39.2 & 53.6 & 35.4 & 51.7 \\
\midrule
\multicolumn{8}{c}{\textbf{\textit{Trained RAG Systems}}} \\
\midrule
\quad Retrobust (Qwen3-4B)~\cite{yoran2024robust} & SFT & 7.2 & 14.7 & 42.4 & 50.0 & 25.2 & 32.4 \\
\quad RAG-DDR (Qwen3-4B)~\cite{li2025ragddr} & RL & 8.2 & 15.3 & 39.8 & 51.9 & 32.8 & 40.8 \\
\quad QAgent$^\dagger$ (Qwen2.5-7B)~\cite{jiang2025qagent} & RL & 7.4 & 13.81 & \underline{42.4} & 46.8 & 35.8 & 36.8 \\
\quad TIRESRAG-R1$^\dagger$ (Qwen2.5-3B-Instruct)~\cite{he2025tiresrag} & RL & \underline{19.4} & \underline{30.0} & 41.0 & \underline{54.2} & \underline{52.8} & \underline{59.6} \\
\midrule\midrule
\multicolumn{8}{c}{\textbf{\textit{Ours}}} \\
\midrule
\quad \textbf{PRISMA} (Qwen3-4B) & RL & \textbf{30.6} & \textbf{39.5} & \textbf{50.2} & \textbf{55.8} & \textbf{57.0} & \textbf{60.1} \\
\bottomrule
\end{tabular*}

\vspace{1em}

\begin{tabular*}{\textwidth}{@{\extracolsep{\fill}}lccccccccccccccc@{}}
\toprule
\textit{Out-of-Distribution Evaluation} & & \multicolumn{2}{c}{\textbf{NQ}~\cite{kwiatkowski2019natural}} & \multicolumn{2}{c}{\textbf{Bamboogle}~\cite{press2023measuring}} & \multicolumn{2}{c}{\textbf{Chemistry}} & \multicolumn{2}{c}{\textbf{Food}} & \multicolumn{2}{c}{\textbf{Game}} & \multicolumn{2}{c}{\textbf{Geography}} & \multicolumn{2}{c}{\textbf{Music}} \\
\textbf{Method} & \textbf{Type} & EM & F1 & EM & F1 & EM & F1 & EM & F1 & EM & F1 & EM & F1 & EM & F1 \\
\midrule
\multicolumn{16}{c}{\textbf{\textit{Closed-book (No Retrieval)}}} \\
\midrule
\quad Qwen3-4B~\cite{qwen3} & Pre-Trained & 3.8 & 14.7 & 2.4 & 7.7 & 49.3 & 53.6 & 23.3 & 31.8 & 23.3 & 31.8 & 21.3 & 26.7 & 44.0 & 48.6 \\
\quad Qwen3-8B~\cite{qwen3} & Pre-Trained & 8.0 & 19.8 & 28.0 & 40.8 & 54.7 & 60.0 & 31.3 & 39.7 & 45.3 & 53.1 & 27.3 & 35.9 & 40.0 & 49.8 \\
\quad Qwen3-30B~\cite{qwen3} & Pre-Trained & 9.2 & 23.1 & 12.8 & 23.2 & 64.0 & 68.1 & 31.3 & 44.1 & 44.0 & 54.2 & 33.3 & 43.2 & 48.7 & 60.8 \\
\quad DeepSeek-V3.2~\cite{liu2025deepseek} & Pre-Trained (API) & 10.8 & 29.0 & 33.6 & 54.4 & 51.3 & 61.5 & 43.3 & 57.7 & 40.0 & 56.7 & 32.7 & 41.2 & 51.3 & 65.7 \\
\quad Gemini-2.5-Flash~\cite{google2025gemini25flash} & Pre-Trained (API) & 16.6 & 34.8 & 50.4 & 63.4 & 66.0 & 71.6 & 40.7 & 56.3 & 56.7 & 70.2 & 40.7 & 50.3 & 56.0 & 69.9 \\
\quad GPT-5-Medium~\cite{openai2025gpt5} & Pre-Trained (API) & 8.8 & 26.6 & 28.0 & 40.1 & 44.0 & 56.8 & 38.0 & 48.6 & 46.7 & 58.5 & 34.7 & 44.3 & 51.3 & 61.7 \\
\midrule
\multicolumn{16}{c}{\textbf{\textit{Single-Step RAG}}} \\
\midrule
\quad Qwen3-4B~\cite{qwen3} & Pre-Trained & 19.2 & 33.5 & 10.4 & 20.6 & 51.3 & 56.1 & 33.3 & 40.4 & 41.3 & 46.4 & 24.7 & 30.2 & 48.0 & 53.0 \\
\quad Qwen3-8B~\cite{qwen3} & Pre-Trained & 16.8 & 31.8 & 20.0 & 26.2 & 44.0 & 49.0 & 24.7 & 29.4 & 36.0 & 41.4 & 23.3 & 27.5 & 34.7 & 41.2 \\
\quad Qwen3-30B~\cite{qwen3} & Pre-Trained & 17.8 & 36.0 & 20.8 & 31.8 & 56.0 & 61.6 & 34.7 & 47.5 & 51.3 & 61.0 & 24.0 & 31.7 & 48.7 & 60.3 \\
\quad DeepSeek-V3.2~\cite{liu2025deepseek} & Pre-Trained (API) & 11.2 & 30.1 & 23.2 & 32.9 & 51.3 & 60.1 & 31.3 & 44.7 & 43.3 & 52.0 & 32.0 & 43.9 & 48.7 & 62.1 \\
\quad Gemini-2.5-Flash~\cite{google2025gemini25flash} & Pre-Trained (API) & 12.0 & 23.4 & 24.0 & 31.8 & 48.7 & 61.6 & 36.0 & 47.2 & 48.0 & 59.3 & 33.3 & 42.4 & 55.3 & 68.9 \\
\quad GPT-5-Medium~\cite{openai2025gpt5} & Pre-Trained (API) & 14.6 & 36.3 & 35.2 & 53.2 & 42.0 & 55.1 & 37.3 & 50.0 & 49.3 & 62.7 & 34.0 & 44.2 & 50.0 & 61.0 \\
\midrule
\multicolumn{16}{c}{\textbf{\textit{Multi-Step RAG$^\ast$ (IRCoT~\cite{trivedi2023interleaving}, RAG-Star~\cite{jiang2025rag}, Search-o1~\cite{li2025search})}}} \\
\midrule
\quad Qwen3-4B~\cite{qwen3} & Pre-Trained & 14.2 & 27.5 & 35.2 & 44.4 & 71.3 & 75.9 & 52.7 & 60.3 & 51.3 & 58.9 & \underline{63.3} & \underline{67.8} & 70.7 & \underline{76.2} \\
\quad Qwen3-8B~\cite{qwen3} & Pre-Trained & 12.4 & 21.2 & 20.8 & 28.9 & 32.7 & 35.8 & 21.3 & 26.9 & 34.0 & 39.8 & 22.0 & 26.5 & 40.7 & 45.9 \\
\quad Qwen3-30B~\cite{qwen3} & Pre-Trained & 17.0 & 33.0 & \underline{45.6} & \underline{58.6} & \underline{74.0} & \underline{77.6} & 52.7 & \underline{64.8} & \underline{57.3} & \underline{69.3} & 57.3 & 63.2 & 67.3 & 74.0 \\
\quad DeepSeek-V3.2~\cite{liu2025deepseek} & Pre-Trained (API) & 18.4 & 34.5 & 58.4 & \uwave{73.9} & 76.0 & 84.3 & 62.0 & 72.2 & 70.7 & 82.0 & 78.7 & \uwave{84.8} & 75.3 & 84.3 \\
\quad Gemini-2.5-Flash~\cite{google2025gemini25flash} & Pre-Trained (API) & \uwave{29.8} & \uwave{47.4} & 59.2 & 69.9 & \uwave{85.3} & \uwave{88.0} & \uwave{72.7} & \uwave{77.7} & \uwave{75.3} & \uwave{82.7} & \uwave{80.0} & 81.6 & \uwave{85.3} & \uwave{88.5} \\
\quad GPT-5-Medium~\cite{openai2025gpt5} & Pre-Trained (API) & 18.0 & 35.4 & \uwave{60.8} & 73.5 & 54.7 & 62.7 & 44.7 & 56.2 & 54.0 & 68.7 & 35.3 & 44.8 & 59.3 & 70.8 \\
\midrule
\multicolumn{16}{c}{\textbf{\textit{Trained RAG Systems}}} \\
\midrule
\quad Retrobust (Qwen3-4B)~\cite{yoran2024robust} & SFT & \underline{27.2} & \underline{39.1} & 24.0 & 34.7 & 56.7 & 58.6 & \underline{53.3} & 56.9 & 50.7 & 56.0 & 40.0 & 41.5 & \underline{72.0} & 73.7 \\
\quad RAG-DDR (Qwen3-4B)~\cite{li2025ragddr} & RL & 20.2 & 31.4 & 32.8 & 44.8 & 60.7 & 69.4 & 42.7 & 54.8 & 52.0 & 63.8 & 46.0 & 50.9 & 56.0 & 69.7 \\
\midrule\midrule
\multicolumn{16}{c}{\textbf{\textit{Ours}}} \\
\midrule
\quad \textbf{PRISMA} (Qwen3-4B) & RL & \textbf{38.6} & \textbf{53.7} & \textbf{46.4} & \textbf{59.2} & \textbf{75.3} & \textbf{78.5} & \textbf{63.3} & \textbf{66.9} & \textbf{73.3} & \textbf{77.1} & \textbf{66.0} & \textbf{68.9} & \textbf{74.7} & \textbf{76.5} \\
\bottomrule
\end{tabular*}

\end{table*}



\paragraph{Baselines}
We evaluate baselines across six backbone LLMs spanning open-source (Qwen3-4B, Qwen3-8B, Qwen3-30B~\cite{qwen3}) and closed-source (DeepSeek-V3.2~\cite{liu2025deepseek}, Gemini-2.5-Flash~\cite{google2025gemini25flash}, GPT-5-Medium~\cite{openai2025gpt5}) models under four pipeline configurations:
\textbf{(1) Closed-book:} Direct LLM inference without retrieval, testing each backbone's parametric knowledge on multi-hop questions.
\textbf{(2) Single-Step RAG:} Single retrieval pass using the question as query, providing top-$k$ passages as context for answer generation.
\textbf{(3) Multi-Step RAG$^\ast$:} Iterative retrieve-and-reason pipelines including IRCoT~\cite{trivedi2023interleaving}, RAG-Star~\cite{jiang2025rag}, and Search-o1~\cite{li2025search}, each performing retrieval conditioned on intermediate reasoning steps.
\textbf{(4) Training-based RAG:} Methods with explicit training on the target task, including Retrobust~\cite{yoran2024robust} (SFT), QAgent~\cite{jiang2025qagent} (RL), TIRESRAG-R1~\cite{he2025tiresrag} (RL), and RAG-DDR~\cite{li2025ragddr} (RL). Retrobust and RAG-DDR use Qwen3-4B, while the cited QAgent results use a Qwen2.5-7B generator and TIRESRAG-R1 uses Qwen2.5-3B-Instruct.

\paragraph{Evaluation Setting Comparison}
As defined in Section~\ref{sec:eval_settings}, evidence provision, corpus scope, and context budgets describe different aspects of an evaluation protocol. Table~\ref{tab:eval_setting_comparison} distinguishes these aspects. Our locally evaluated retrieval baselines use the same 21M-passage corpus and retrieval infrastructure.

\paragraph{Metrics}
We report \textbf{Exact Match (EM)} and \textbf{F1} following standard practice~\cite{yang2018hotpotqa}.

\paragraph{Implementation Details}
Locally evaluated retrieval baselines use identical retrieval infrastructure with FAISS~\cite{johnson2019billion} and BGE-M3~\cite{chen2024bge}. Our local retrieval experiments use the \textbf{full} Wikipedia corpus from DPR~\cite{karpukhin2020dense} (\textbf{21M} passages). Retrieval and cache parameters are reported in Table~\ref{tab:retrieval_config}. All experiments use 4 NVIDIA H100 GPUs.

\subsection{Main Results}
\label{sec:main_results}

\paragraph{PRISMA outperforms trained systems and reaches API-comparable performance}
As shown in Table~\ref{tab:main_results}, \textbf{PRISMA} consistently outperforms prior trained RAG systems on both in-distribution and out-of-distribution benchmarks. On in-distribution, PRISMA achieves 30.6\%/39.5\% on MuSiQue, 50.2\%/55.8\% on HotpotQA, and 57.0\%/60.1\% on 2WikiMHQA, surpassing the strongest baseline TIRESRAG-R1 by +11.2/+9.5 points on MuSiQue. For out-of-distribution evaluation, PRISMA leads all trained RAG baselines across all seven benchmarks, notably on NQ (38.6\%/53.7\%) and Bamboogle (46.4\%/59.2\%). It also surpasses GPT-5-Medium on MuSiQue (+7.2/+4.5 points) and achieves the best overall NQ result, exceeding the strongest closed-source API configuration by +8.8/+6.3 points. Taken together across all ten benchmarks, these results show that a 4B open-source system can reach an overall performance level comparable to larger closed-source API models.

\paragraph{Closed-source API models benefit from scale but lack robustness}
With multi-step retrieval, closed-source API models perform well when sufficient intermediate evidence is available. For example, Gemini-2.5-Flash reaches 29.8\%/47.4\% on NQ and 85.3\%/88.0\% on Chemistry, while DeepSeek-V3.2 attains 78.7\%/84.8\% on Geography and 75.3\%/84.3\% on Music (Table~\ref{tab:main_results}). However, performance drops significantly when bridge evidence is missing or retrieval quality fluctuates, highlighting the need for explicit planning and in-loop verification rather than relying solely on model priors.

\paragraph{Additional observations}
\textbf{(i) Single-step RAG can hurt.} Unfiltered single-step RAG often underperforms closed-book prompting. For example, on HotpotQA, Gemini-2.5-Flash drops from 38.4\%/51.4\% to 19.2\%/29.9\% (Table~\ref{tab:main_results}), emphasizing the need for retrieval quality control.
\textbf{(ii) Small expert models can be competitive.} Qwen3-4B and Qwen3-30B perform similarly on out-of-distribution domains (e.g., Chemistry), showing that smaller, well-aligned models can match larger ones in specific tasks, supporting cost-effective deployment.

\paragraph{PRISMA generalizes robustly to OOD while existing trained methods overfit}
Existing RL-trained methods show strong in-distribution performance but degrade on OOD benchmarks. On domain-specific datasets, zero-shot Multi-Step RAG with Qwen3-4B often matches or outperforms the trained Retrobust and RAG-DDR despite having no task-specific optimization. It exceeds both methods on Chemistry (71.3\% vs.\ 56.7\%/60.7\%) and Geography (63.3\% vs.\ 40.0\%/46.0\%); on Game and Music, its EM is within 1.3 points of the stronger trained baseline. QAgent and TIRESRAG-R1 are not included in the seven-benchmark OOD comparison reported here. This pattern suggests that end-to-end RL training can specialize to dataset-specific heuristics. In contrast, PRISMA maintains strong OOD performance (e.g., Chemistry: 75.3\%, Game: 73.3\%) by decoupling agent training: schema-defined responsibility boundaries reduce cross-module overfitting and enable each agent's learned skills to transfer across domains.

\subsection{Ablation Study}
\label{sec:ablation}

We ablate on the MuSiQue evaluation set to assess each module's contribution and the impact of our training strategy. Results are in Table~\ref{tab:ablation_combined}.

\begin{table}[!t]
\caption{Component and Training Ablations on the MuSiQue Evaluation Set}
\label{tab:ablation_combined}
\centering
\small
\setlength{\tabcolsep}{2pt}
\begin{tabular*}{\columnwidth}{@{\extracolsep{\fill}}lccc}
\toprule
\textbf{Configuration} & \textbf{EM} & \textbf{F1} & \textbf{$\Delta$EM} \\
\midrule
\multicolumn{4}{c}{\textbf{\textit{Component Ablation}}} \\
\midrule
\textbf{PRISMA} & \textbf{30.6} & \textbf{39.5} & -- \\
w/o Planner & 15.8 & 23.3 & $-$14.8 \\
w/o Inspector & 22.4 & 29.0 & $-$8.2 \\
w/o Context Inspector & 24.6 & 31.6 & $-$6.0 \\
w/o Reasoning Inspector & 23.4 & 31.0 & $-$7.2 \\
\midrule
\multicolumn{4}{c}{\textbf{\textit{Training Ablation}}} \\
\midrule
\textbf{PRISMA} & \textbf{30.6} & \textbf{39.5} & -- \\
Planner $\rightarrow$ Base Model & 24.4 & 32.9 & $-$6.2 \\
Solver $\rightarrow$ Base Model & 24.0 & 32.5 & $-$6.6 \\
Inspector $\rightarrow$ Base Model & 25.4 & 33.1 & $-$5.2 \\
All Agents $\rightarrow$ Base Model & 20.0 & 30.8 & $-$10.6 \\
\bottomrule
\end{tabular*}
\end{table}
\begin{table}[!t]
\caption{Success Pattern Distribution (153 Correct MuSiQue Answers)}
\label{tab:success_patterns}
\centering
\small
\begin{tabular}{@{}p{4.8cm}rr@{}}
\toprule
\textbf{Success Pattern} & \textbf{Count} & \textbf{\%} \\
\midrule
Clean execution (no intervention) & 50 & 32.7 \\
\emph{Context Inspector} only & 36 & 23.5 \\
\emph{Reasoning Inspector} only & 26 & 17.0 \\
Both inspectors & 41 & 26.8 \\
\bottomrule
\end{tabular}
\end{table}




\subsubsection{Component Ablation}
We ablate each component in turn, keeping the rest of the pipeline unchanged.
Removing the Planner and falling back to single-query retrieval leads to the largest drop (EM: 30.6 $\rightarrow$ 15.8, $\Delta$EM = $-14.8$), confirming the necessity of dependency-aware decomposition.
Removing the Inspector entirely degrades EM to 22.4 ($\Delta$EM = $-8.2$), highlighting the value of in-loop verification and repair. Removing the Context Inspector yields a moderate drop ($\Delta$EM = $-6.0$), while removing the Reasoning Inspector leads to a larger degradation ($\Delta$EM = $-7.2$), indicating that post-execution verification is most critical.

\subsubsection{Training Ablation}
We ablate task-specific optimization by replacing each trained agent with its base model.
Replacing the Solver yields the largest single-agent drop ($\Delta$EM = $-6.6$), indicating that evidence-grounded multi-document synthesis is a primary bottleneck. Replacing the Planner ($\Delta$EM = $-6.2$) and Inspector ($\Delta$EM = $-5.2$) also degrade results. When all agents are replaced, performance drops sharply ($\Delta$EM = $-10.6$), confirming that PRISMA's gains are distributed across agents rather than driven by any single module. This validates the decoupled training strategy: each agent's schema-defined specialization contributes meaningfully to end-to-end performance.

\subsubsection{Empirical Justification of Modular Design}
The ablation results provide empirical support for the modular architecture. The Planner (decomposition) and Inspector (verification) contribute $\Delta$EM of $-14.8$ and $-8.2$ respectively, confirming that these stages are individually necessary. Inspector-triggered interventions occur in 67.3\% of correct answers (Table~\ref{tab:success_patterns}), showing that interleaved verification is frequently activated during successful execution. The distributed drops across all agents validate the schema-defined responsibility partitioning and agent-specific reward design.

\begin{figure*}[!t]
\centering
\includegraphics[width=\textwidth]{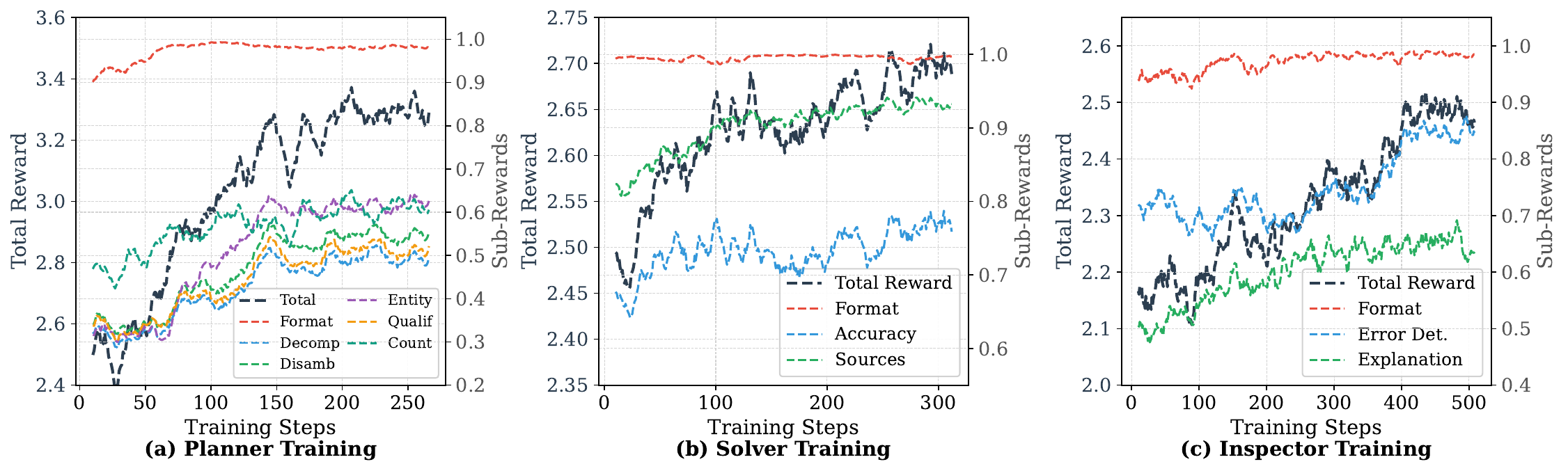}
\caption{GRPO training dynamics for the three specialized agents. (a) Planner training tracks total reward together with format, decomposition, disambiguation, entity, qualification, and hop-count rewards. (b) Solver training tracks total reward, format compliance, answer accuracy, and source-grounding rewards. (c) Inspector training tracks total reward, format compliance, error detection, and explanation quality. In all panels, total rewards increase and then plateau, indicating stable convergence under decoupled, role-specific optimization.}
\label{fig:reward_curves}
\end{figure*}
\subsection{Diagnostic Analysis}
\label{sec:diagnostic_analysis}

\subsubsection{Performance by Question Complexity}
\label{sec:hop_analysis}
\begin{table}[!t]
\caption{EM by Hop Count on the MuSiQue Evaluation Set. $^\ast$Best of IRCoT, RAG-Star, Search-o1}
\label{tab:hop_analysis}
\centering
\small
\begin{tabular*}{\columnwidth}{@{\extracolsep{\fill}}lccc@{}}
\toprule
\textbf{Method} & \textbf{2-hop} & \textbf{3-hop} & \textbf{4-hop} \\
\midrule
Single-Step RAG & 14.7 & 0.7 & 2.2 \\
Multi-Step RAG$^\ast$ & 11.3 & 4.8 & 1.1 \\
\textsc{PRISMA} w/ Base & 32.1 & 8.3 & 3.3 \\ 
\textsc{PRISMA} (RL) & \textbf{41.1} & \textbf{26.9} & 5.6 \\
\bottomrule
\end{tabular*}
\end{table}
\begin{table}[!t]
\caption{Retrieval Recall Breakdown on MuSiQue}
\label{tab:retrieval_analysis}
\centering
\small
\begin{tabular*}{\columnwidth}{@{\extracolsep{\fill}}p{0.55\columnwidth}cc@{}}
\toprule
\textbf{Configuration} & \textbf{Recall} & \textbf{Improv.} \\
\midrule
Single-query baseline & 37.1 & -- \\
+ Decomposition & 45.8 & +8.7 \\
+ Context Inspector + Expansion & 50.8 & +5.0 \\
\midrule
\textbf{Total} & \textbf{50.8} & \textbf{+13.7} \\
\bottomrule
\end{tabular*}
\end{table}
\begin{table}[!t]
\caption{Direct Impact of Inspector-Triggered Recovery on the 500-Question MuSiQue Evaluation Set}
\label{tab:inspection_impact}
\centering
\small
\begin{tabular*}{\columnwidth}{@{\extracolsep{\fill}}lcc@{}}
\toprule
\textbf{Measure} & \textbf{Result} & \textbf{Change/Share} \\
\midrule
EM before $\rightarrow$ after & 28.4 $\rightarrow$ 30.6 & +2.2 \\
F1 before $\rightarrow$ after & 36.0 $\rightarrow$ 39.5 & +3.4 \\
Improved ($0\rightarrow1$ EM) & 13 & 2.6\% \\
Regressed ($1\rightarrow0$ EM) & 2 & 0.4\% \\
Unchanged & 485 & 97.0\% \\
\bottomrule
\end{tabular*}
\end{table}


Table~\ref{tab:hop_analysis} stratifies MuSiQue results by hop count. RL-trained \textsc{PRISMA} achieves the strongest performance on 2-hop questions (41.1\% EM) and 3-hop questions (26.9\% EM), outperforming the best multi-step RAG baseline by 29.8 and 22.1 points, respectively. These gains indicate that specialized Planner and Solver training improves decomposition and multi-document synthesis where most multi-hop questions lie.

Performance drops for all systems on 4-hop questions, reflecting the compounded difficulty of long retrieval-reasoning chains. \textsc{PRISMA} still obtains the highest 4-hop EM among the compared systems (5.6\%), while the gap between the RL-trained and base variants is small at this sample size: 5.6\% versus 3.3\%, corresponding to two questions. This suggests that the remaining bottleneck is dominated by retrieval-chain failure rather than insufficient agent specialization alone. Formally, if $R_i$ denotes successful evidence retrieval at hop $i$, complete-chain success requires $P(R_1)\prod_{i=2}^{4}P(R_i\mid R_1,\ldots,R_{i-1})$; under a simplified shared-recall approximation, this scales as $r^4$. Thus, Inspector-guided expansion can mitigate but not eliminate the long-chain retrieval bottleneck.

Finally, replacing all trained agents with the base model substantially reduces 2-hop and 3-hop performance (41.1\% $\rightarrow$ 32.1\% and 26.9\% $\rightarrow$ 8.3\%), confirming that \textsc{PRISMA}'s main gains come from specialized agent training rather than the backbone alone.

\subsubsection{Retrieval Effectiveness Analysis}
\label{sec:retrieval_analysis}


\noindent\textbf{Retrieval System Effectiveness.} Table~\ref{tab:retrieval_analysis} shows the incremental improvement from each component on MuSiQue. Decomposition contributes +8.7\% recall by converting multi-hop queries into focused single-hop subquestions. The Context Inspector with retrieval expansion adds another +5.0\%, demonstrating the value of in-loop verification for retrieval quality.

\subsubsection{Inspector Impact and Error Analysis}
\label{sec:qualitative}


\noindent\textbf{Direct Before--After Impact.} Table~\ref{tab:inspection_impact} directly compares the original Solver answer before Inspector-triggered recovery with the final answer after inspection on the same 500-question MuSiQue run used for the main result. Inspection increases EM from 28.4\% to 30.6\% and F1 from 36.0\% to 39.5\%. At the question level, 13 cases improve from incorrect to correct, while only two regress from correct to incorrect, yielding a net gain of 11 correct answers and a recovery-to-regression ratio of 6.5:1. The remaining 485 cases are unchanged, indicating a positive net effect without widespread over-intervention.

\noindent\textbf{Success Pattern Analysis.} We analyze the 153 correctly answered questions (EM=1) in the MuSiQue evaluation set from the same run used for the main result. Table~\ref{tab:success_patterns} shows that Inspector-triggered interventions occur in 67.3\% of successful cases. The Context Inspector is involved in 50.3\% of successes (23.5\% alone and 26.8\% together with the Reasoning Inspector), while the Reasoning Inspector is involved in 43.8\% (17.0\% alone and 26.8\% together). These are intervention frequencies among successful cases rather than causal estimates of the Inspectors' contribution.

\begin{table}[!t]
\caption{Failure Mode Distribution (347 Failed MuSiQue Cases)}
\label{tab:failure_modes}
\centering
\scriptsize
\begin{tabular*}{\columnwidth}{@{\extracolsep{\fill}}p{0.58\columnwidth}rr@{}}
\toprule
\textbf{Failure Mode} & \textbf{Count} & \textbf{\%} \\
\midrule
\multicolumn{3}{c}{\textbf{\textit{Detected Error Events (2,555 total)}}} \\
\midrule
Retrieval errors & 1650 & 64.6 \\
Reasoning errors & 834 & 32.6 \\
Subquestion errors (Planner) & 39 & 1.5 \\
Extraction errors (Solver) & 32 & 1.3 \\
\midrule
\multicolumn{3}{c}{\textbf{\textit{Detected-Event Categories}}} \\
\midrule
Retrieval-related (retr.\ + subq.) & 1689 & 66.1 \\
Reasoning-related (reas.\ + extr.) & 866 & 33.9 \\
\midrule
\multicolumn{3}{c}{\textbf{\textit{Failure-Case Detection (347 questions)}}} \\
\midrule
At least one error flagged & 311 & 89.6 \\
No error flagged & 36 & 10.4 \\
\bottomrule
\end{tabular*}
\end{table}

\noindent\textbf{Failure Mode Analysis.} Table~\ref{tab:failure_modes} summarizes the detected error events among the 347 failed questions. Retrieval-related events account for 66.1\% and reasoning-related events for 33.9\%. Because a question can generate multiple Inspector events across hops and retries, these percentages describe event composition rather than shares of failed questions. At the question level, at least one error is flagged in 311 of 347 failures (89.6\%), while 36 failures (10.4\%) receive no error flag. Together with the success-side intervention frequencies in Table~\ref{tab:success_patterns}, these results characterize how often the Inspectors activate and what errors they report without treating activation as a causal estimate of recovery.

\subsection{Training Analysis}
\label{sec:training_analysis}

\noindent\textbf{GRPO training yields stable and role-specific optimization.} Fig.~\ref{fig:reward_curves} shows stable GRPO training for all three agents, with total reward increasing smoothly and then plateauing. For the Planner, format reward saturates early while semantic sub-rewards (e.g., decomposition/disambiguation) keep improving. For the Solver, accuracy and citation faithfulness improve in tandem. For the Inspector, detection accuracy dominates the reward signal.

\subsection{Efficiency Analysis}
\label{sec:efficiency_analysis}

\noindent\textbf{Efficiency analysis isolates retrieval, caching, and end-to-end runtime costs.} All efficiency experiments follow the main evaluation setting and report wall-clock latency per question. We evaluate three aspects: retrieval-configuration cost on MuSiQue, Memoize caching over a MuSiQue evaluation set of 1,000 questions, and component-level latency for a representative three-hop execution. For Memoize, the cache starts empty, is shared across the evaluation run, populated online, and cleared after the run, so later questions can reuse semantically matched subquestions accumulated earlier in the same run.

\subsubsection{Retrieval Configuration Efficiency}

We evaluate retrieval configurations separately to examine accuracy--efficiency trade-offs. Table~\ref{tab:retriever_ablation} shows that the full three-stage cascade (100$\rightarrow$30$\rightarrow$10) achieves the best recall (59.6\%) and EM (33.0\%) while maintaining competitive latency (20.8s), validating the cascade design over simpler alternatives.

\begin{table}[!t]
\caption{Retrieval Configuration Ablation on the MuSiQue Evaluation Set}
\label{tab:retriever_ablation}
\centering
\scriptsize
\begin{tabular*}{\columnwidth}{@{\extracolsep{\fill}}lccc@{}}
\toprule
\textbf{Configuration} & \textbf{Recall} & \textbf{Lat.} & \textbf{EM} \\
\midrule
Dense only (k=10) & 57.5 & 71.0s & 30.0 \\
Dense + Sparse (30$\rightarrow$10) & 58.5 & 21.4s & 31.0 \\
Dense + Rerank (100$\rightarrow$10) & 56.1 & 71.8s & 28.0 \\
Full (100$\rightarrow$30$\rightarrow$10) & 59.6 & 20.8s & 33.0 \\
\bottomrule
\end{tabular*}
\end{table}

\subsubsection{Memoize Efficiency}

\begin{figure}[!t]
\centering
\includegraphics[width=\columnwidth]{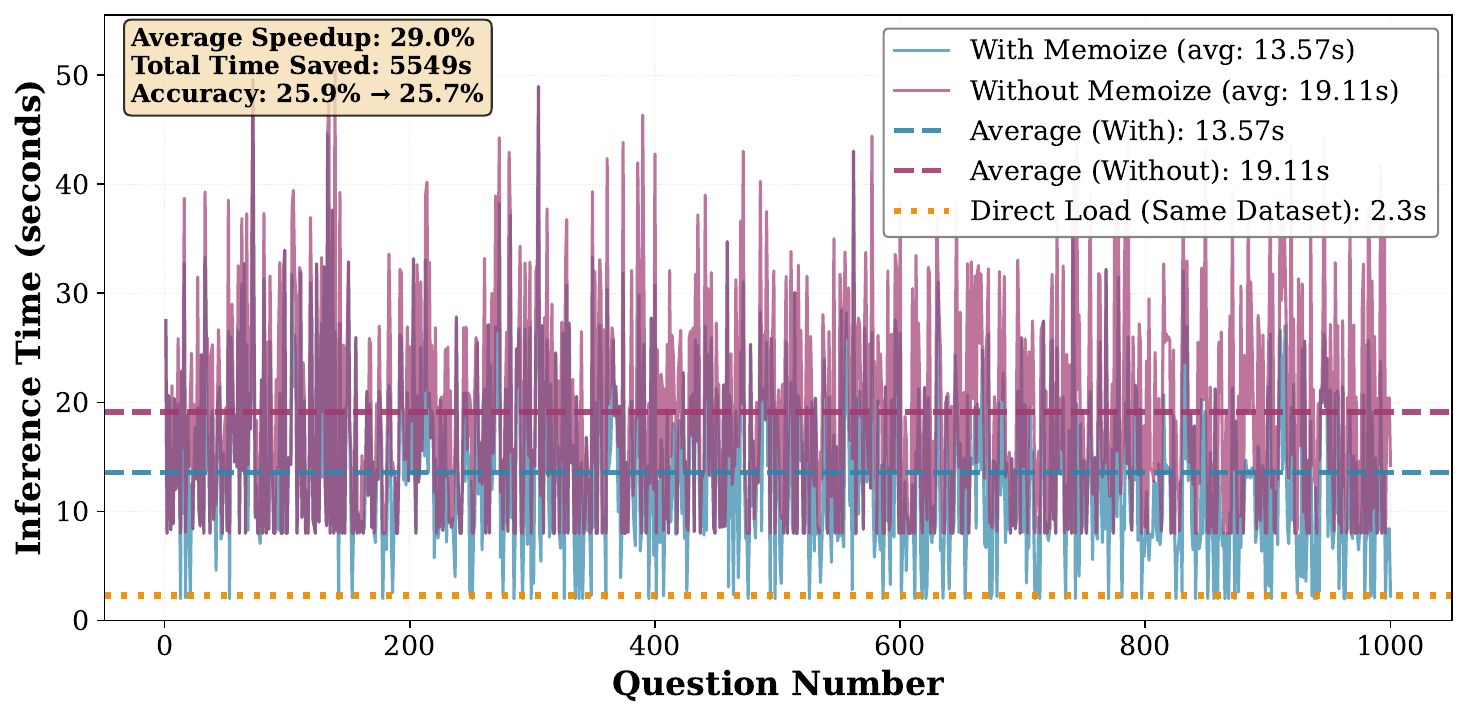}
\caption{Memoize efficiency on a MuSiQue evaluation set of 1000 questions. Blue: with caching (avg 13.57s); purple: without caching (avg 19.11s); orange: theoretical minimum (2.3s).}
\label{fig:memoize_efficiency}
\end{figure}

Fig.~\ref{fig:memoize_efficiency} evaluates PRISMA's semantic memory cache on 1000 MuSiQue questions. Enabling Memoize reduces average inference time from 19.11s to 13.57s per question (a 29.0\% reduction in latency), saving 92.5 minutes over 1000 questions, with negligible accuracy impact ($\Delta$EM = $-0.2$). Because the cache starts empty and is populated online, its benefit grows as the run progresses: later questions can bypass the Retrieve--Inspect--Solve loop when a semantically equivalent subquestion has already been answered. The remaining gap to the 2.3s theoretical minimum indicates that many subquestions still require fresh retrieval, consistent with MuSiQue's diverse compositional queries.

\subsubsection{End-to-End Latency Breakdown}

\begin{figure}[!t]
\centering
\includegraphics[width=\columnwidth]{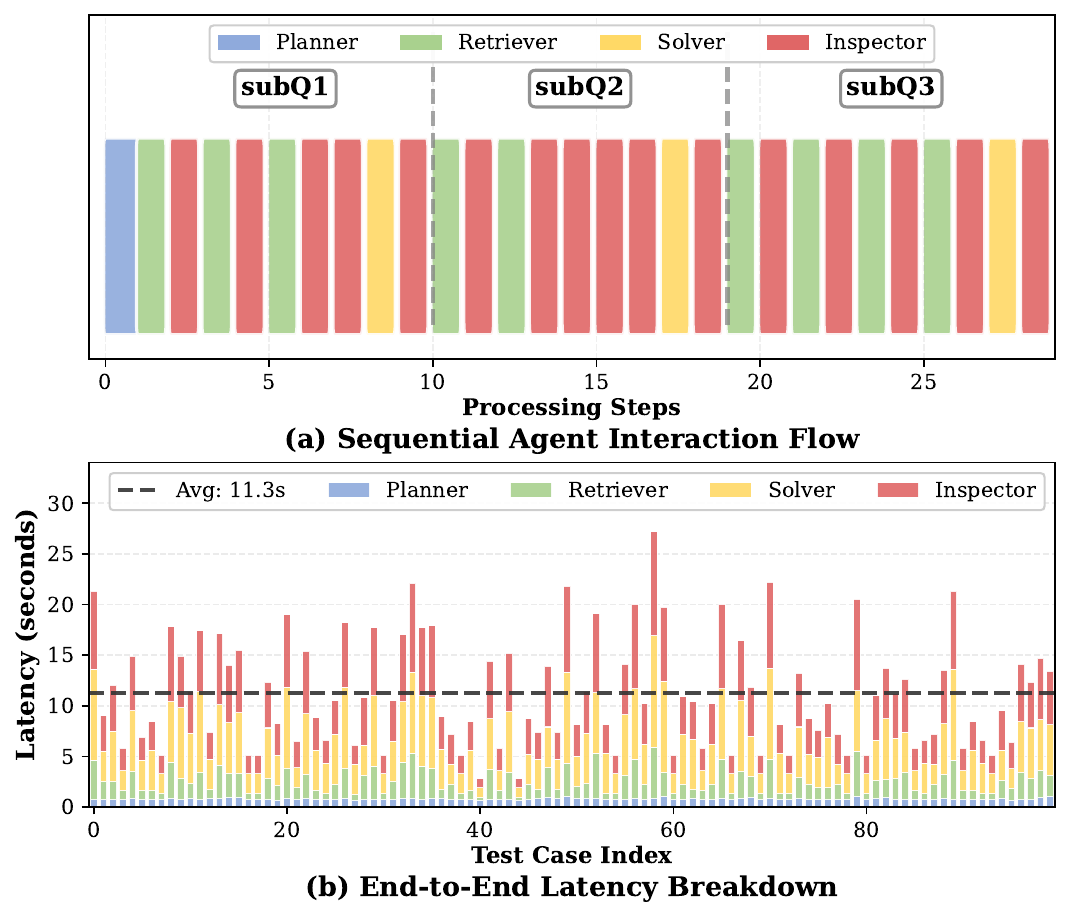}
\caption{\textbf{Top:} Agent interaction flow for a sample three-hop question. \textbf{Bottom:} Latency breakdown per component.}
\label{fig:system_analysis}
\end{figure}

Fig.~\ref{fig:system_analysis} shows the end-to-end latency breakdown for a typical three-hop chain. Inspection and solving dominate the runtime, with planning and retrieval contributing less. This aligns with PRISMA's design to prioritize verification and grounding, while the Memoize results above show how within-run reuse reduces redundant computation.

\subsection{Case Studies}
\label{sec:case_studies}

We present two condensed real traces from the same 500-question MuSiQue run to illustrate both the benefit and the boundary of PRISMA's dual-inspection mechanism. Table~\ref{tab:case_success} shows a successful targeted recovery: the Context Inspector correctly accepts sufficient evidence, while the Reasoning Inspector catches a same-name entity-binding error and triggers a repair that recovers the gold answer. Table~\ref{tab:case_failure} shows a complementary failure with safe abstention: the plan, entity binding, and recovery routing remain correct, but repeated retrieval fails to recall the decisive numeric passage. Together, the cases show that schema-defined inspection localizes errors and enables recovery when evidence is available, while preventing unsupported fabrication when the open-domain corpus lookup remains incomplete.


\begin{table}[t!]
\caption{Condensed Real Trace: Recovery from Same-Name Entity Confusion}
\label{tab:case_success}
\centering
\scriptsize
\begin{tabular}{|p{0.95\columnwidth}|}
\hline
\textbf{Trace ID:} \texttt{2hop\_\_799936\_20273} (MuSiQue) \\
\hline
\textbf{Question:} What year did the author of \emph{The First Blast of the Trumpet Against the Monstruous Regiment of Women} die? \\
\textbf{Gold answer:} 1572 \\
\hline
\textbf{Planner:} (1) identify the work's author; (2) ask when \texttt{[Answer1]} died. \\
\hline
\textbf{Hop 1:} Document [2] identifies the author as the Scottish reformer John Knox. Both inspectors return \texttt{none}; the Solver returns \textbf{John Knox}. \\
\hline
\textbf{Hop 2 evidence:} The retrieved set contains several people named John Knox. Documents [3] and [4] identify the Scottish Reformation leader and state that he died in 1572, whereas [5] describes a philanthropist who died in 1790. \\
\hline
\textbf{Context Inspector:} \texttt{error\_stage=none}. It correctly notes that [3] and [4] provide sufficient evidence for the author identified at Hop 1. \\
\hline
\textbf{Solver before post-solution inspection:} \textbf{1790} \texttimes. It selects the philanthropist rather than the author. \\
\hline
\textbf{Reasoning Inspector:} \texttt{error\_stage=reasoning}. It identifies the entity-binding error and points to [3] and [4], which support 1572 for the Scottish reformer. \\
\hline
\textbf{Targeted repair:} The diagnosis yields the query ``What year did John Knox, the author of \emph{The First Blast of the Trumpet Against the Monstruous Regiment of Women}, die?'' One post hoc retrieval round is executed. \\
\hline
\textbf{Solver after repair:} \textbf{1572}; the Reasoning Inspector re-check returns \texttt{none}. \\
\hline
\textbf{Final:} \textbf{1572} \checkmark \quad (EM: $0\!\rightarrow\!1$; F1: $0\!\rightarrow\!1$) \\
\hline
\end{tabular}
\end{table}

\begin{table}[t!]
\caption{Condensed Real Failure Trace: Safe Abstention after Exhaustive Retrieval Repair}
\label{tab:case_failure}
\centering
\scriptsize
\begin{tabular}{|p{0.95\columnwidth}|}
\hline
\textbf{Trace ID:} \texttt{2hop\_\_813461\_31270} (MuSiQue) \\
\hline
\textbf{Question:} How many miles are there between Nashville and the place in Tennessee where Mark Day was born? \\
\textbf{Gold answer:} 45 \\
\hline
\textbf{Benchmark support chain:} Mark Day was born in Clarksville, Tennessee; the annotated supporting passage states that Clarksville is about 45 miles northwest of Nashville. \\
\hline
\textbf{Planner:} (1) identify Mark Day's Tennessee birthplace; (2) ask the distance between Nashville and \texttt{[Answer1]}. \\
\hline
\textbf{Hop 1:} Retrieved evidence explicitly states that Mark Day was born in Clarksville. Both inspectors return \texttt{none}; the Solver returns \textbf{Clarksville}. \\
\hline
\textbf{Hop 2 initial retrieval:} The documents mention routes connecting Clarksville and Nashville, but none states the distance in miles. \\
\hline
\textbf{Context Inspector:} \texttt{error\_stage=retrieval}. It identifies the missing numeric attribute and rewrites the query as ``How many miles are there between Nashville and Clarksville, Tennessee?'' Three expansion rounds still fail to retrieve the 45-mile statement. \\
\hline
\textbf{Solver:} \textbf{Not found in the documents}. The Reasoning Inspector escalates the unresolved refusal, triggering two post hoc retrieval attempts with the query ``distance in miles between Nashville and Clarksville Tennessee.'' Neither attempt retrieves an explicit mileage. \\
\hline
\textbf{Final:} \textbf{Not found in the documents} \texttimes \quad (EM $=0$; F1 $=0$) \\
\hline
\textbf{Attribution and limitation:} The plan, entity binding, and recovery routing remain correct, but seven total retrieval calls (including Hop 1) never recall the decisive numeric passage. The schema turns this into a traceable retrieval-recall failure, while evidence-grounded abstention prevents hallucination. \\
\hline
\end{tabular}
\end{table}
\section{Related Work}

\subsection{Retrieval-Augmented Generation}
RAG grounds LLM generation in external evidence~\cite{lewis2020retrieval}, with retrieval improved by dense passage retrieval~\cite{karpukhin2020dense}, hybrid sparse-dense methods~\cite{ma2021zero,luan2021sparse}, and adaptive retrieval strategies~\cite{asai2024selfrag,jeong2024adaptiverag}. Recent advances include rationale denoising~\cite{wei2025instructrag}, speculative drafting~\cite{wang2025speculative}, and planning-aware pipelines~\cite{lee2024planrag,liu2025opera}. For multi-hop queries, iterative retrieval frameworks such as ReAct~\cite{yao2023react}, IRCoT~\cite{trivedi2023interleaving}, RAG-Star~\cite{jiang2025rag}, Search-o1~\cite{li2025search}, and DualRAG~\cite{chen2024dualrag} interleave retrieval with reasoning to progressively gather evidence. However, retrieval and reasoning remain loosely coupled without schema-defined coordination: missing critical passages at early hops collapses downstream reasoning with limited recovery.

\subsection{Multi-Hop Question Answering}
Multi-hop QA is evaluated on HotpotQA~\cite{yang2018hotpotqa}, 2WikiMultiHopQA~\cite{ho2020constructing}, and MuSiQue~\cite{trivedi2022musique}, requiring evidence synthesis across multiple passages. A key distinction is whether evidence is supplied with the question or must be retrieved. ARENA~\cite{ren2025arena} operates on provided paragraphs. Methods that retrieve from a corpus include MMOA~\cite{chen2025mmoa}, which selects passages retrieved from Wikipedia, and OPERA~\cite{liu2025opera}, which uses planning-aware retrieval. Chain-of-Agents~\cite{zhang2024chainagents} processes segmented portions of a provided long context. In open-domain QA, retrieval must identify relevant evidence from the available corpus, where ambiguous queries can reduce recall~\cite{xiong2024beam,tang2024multihop}. Advances in bridge reasoning~\cite{ke2024bridging} and RL-based optimization~\cite{he2025tiresrag,yoran2024robust} have improved specific capabilities, but challenges persist in disconnected retrieval-reasoning pipelines, bridge answer failures, and the absence of in-loop verification over evidence chains. Evidence provision, corpus scope, and context budgets should therefore be reported explicitly when comparing methods.

\subsection{LLM-based Multi-Agent Systems}
LLM-based multi-agent systems distribute complex tasks across specialized agents with tool-augmented decomposition~\cite{schick2023toolformer,qin2023toolllm}, multi-path reasoning~\cite{yao2023tree,besta2024graph}, and collaborative coordination~\cite{hong2023metagpt,wu2023autogen}, applied to code tasks~\cite{jimenez2024swebench} and large-scale coordination~\cite{wang2025megagent}. For RAG, recent multi-agent pipelines such as MMOA~\cite{chen2025mmoa}, TIRESRAG-R1~\cite{he2025tiresrag}, and QAgent~\cite{jiang2025qagent} attempt to distribute retrieval and reasoning across agents. However, a recurring gap is the lack of effective inter-agent communication protocols: most systems define agent roles but not the structured interfaces through which agents exchange information and trigger corrective actions. Without such schema-defined coordination, error detection and targeted recovery remain ad hoc.

\subsection{Reinforcement Learning for LLMs}
Reinforcement learning has been widely adopted to align LLM behavior, from RLHF with PPO~\cite{schulman2017proximal,ouyang2022training} to critic-free alternatives such as DPO~\cite{rafailov2023direct} and GRPO~\cite{shao2024deepseekmath}. PPO incurs substantial critic overhead~\cite{santacroce2023efficient} and suffers from high-variance updates~\cite{dong2023rlhf} and reward hacking~\cite{casper2024reward}, while DPO can lag on complex reasoning~\cite{xu2024dpo}. GRPO uses group-based advantage estimation to reduce compute and variance, making it suitable for multi-agent pipelines where each agent requires independent optimization. In the RAG domain, RL has been applied to optimize partial skills such as retrieval selection~\cite{asai2024selfrag,li2025ragddr} and reasoning quality~\cite{ren2025arena}, but holistic optimization across multi-agent pipelines with proper credit attribution remains underexplored.


\section{Conclusion}
We present \textsc{PRISMA}, a multi-agent framework for open-domain multi-hop question answering that coordinates agents through an inter-agent communication schema. Each agent's structured output serves as both its own reasoning constraint and the downstream agent's input interface, enabling modular error attribution and targeted recovery via dual-phase Inspector verification. A decoupled training strategy separates agent specialization from trajectory-conditioned verification on a frozen pipeline. Across all ten benchmarks over a 21M-passage corpus, PRISMA consistently surpasses the reported trained baselines and delivers overall performance competitive with larger closed-source API models, while maintaining consistent out-of-distribution generalization where existing RL-trained methods degrade. PRISMA demonstrates that schema-coordinated multi-agent architectures provide an effective paradigm for dynamic interacting systems in knowledge-intensive retrieval.

\bibliographystyle{IEEEtran}
\bibliography{sample}
\bstctlcite{IEEEexample:BSTcontrol}

\clearpage
\appendices

\section{Training Configuration}
\label{sec:training_config}

All agents are fine-tuned from the same base model (Qwen3-4B-Instruct) using full-parameter training with GRPO. Table~\ref{tab:training_config} summarizes the infrastructure, and Table~\ref{tab:merged_training_config} reports per-agent dataset sizes and optimization budgets.

\begin{table}[h]
\caption{Infrastructure Configuration}
\label{tab:training_config}
\centering
\scriptsize
\begin{tabular*}{\columnwidth}{@{\extracolsep{\fill}}lccc@{}}
\toprule
\textbf{Component/Method} & \textbf{Training} & \textbf{Hardware} & \textbf{Inference} \\
\midrule
\multicolumn{4}{c}{\textbf{\textit{\textsc{PRISMA} Components (Decoupled Training)}}} \\
\midrule
Planner & GRPO & 4$\times$ H100 & vLLM \\
Solver & GRPO & 4$\times$ H100 & vLLM \\
Inspector (unified) & GRPO & 4$\times$ H100 & vLLM \\
\midrule
\multicolumn{4}{c}{\textbf{\textit{Baselines}}} \\
\midrule
Retrobust & SFT & 4$\times$ H100 & vLLM \\
RAG-DDR & DPO & 4$\times$ H100 & vLLM \\
IRCoT / RAG-Star / Search-o1 & Zero-shot & -- & Greedy (temp=0) \\
DeepSeek-V3.2 & Zero-shot & -- & API (temp=0) \\
Gemini-2.5-Flash & Zero-shot & -- & API (temp=0) \\
GPT-5 & Zero-shot & -- & API (temp=0) \\
\bottomrule
\end{tabular*}
\end{table}

\begin{table}[h]
\caption{GRPO Hyperparameters}
\label{tab:merged_training_config}
\centering
\scriptsize
\resizebox{\columnwidth}{!}{%
\begin{tabular}{@{}lccc@{}}
\toprule
\textbf{Parameter} & \textbf{Planner} & \textbf{Solver} & \textbf{Inspector} \\
\midrule
\multicolumn{4}{c}{\textbf{\textit{Agent-Specific GRPO Hyperparameters}}} \\
\midrule
Dataset size (prompts) & 8,563 & 20,000 & 8,141 \\
Training epochs & 2 & 1 & 2 \\
Number of GPUs & 4 & 4 & 4 \\
Per-device batch size & 4 & 4 & 4 \\
Gradient accumulation steps & 32 & 32 & 16 \\
\midrule
Sampled generations (per prompt) & 8 & 8 & 16 \\
Generations used (per prompt) & 8 & 8 & 8 \\
Global batch size & 512 & 512 & 256 \\
Token-loss aggregation & Sequence mean & Sequence mean & Global-token mean \\
KL coefficient $\beta$ & 0.04 & 0.04 & 0.0 \\
\midrule
Effective train samples & 68,504 & 160,000 & 65,128 \\
Total training steps & 266 & 312 & 508 \\
\midrule
\multicolumn{4}{c}{\textbf{\textit{Shared Configuration}}} \\
\midrule
Learning rate & \multicolumn{3}{c}{3.0e-6} \\
Clipping coefficient $\varepsilon_{\mathrm{c}}$ & \multicolumn{3}{c}{0.2} \\
Temperature / top-k / top-p & \multicolumn{3}{c}{1.0 / 20 / 0.9} \\
Base model & \multicolumn{3}{c}{Qwen3-4B-Instruct} \\
Corpus & \multicolumn{3}{c}{DPR Wikipedia (21M passages)} \\
Retriever / Reranker & \multicolumn{3}{c}{BGE-M3 / BGE-reranker-v2-m3} \\
Planner judge model & \multicolumn{3}{c}{Qwen3-30B-A3B-Instruct-2507} \\
Audit teacher $g_T$ & \multicolumn{3}{c}{Qwen3-Max} \\
\bottomrule
\end{tabular}%
}
\end{table}

\paragraph{Reward Implementation Details}
For the Planner reward in Eq.~\eqref{eq:r_plan_main}, we set $w_{\mathrm{fmt}}^{P}=1.0$, $w_{\mathrm{cnt}}^{P}=2.0$, and $w_{\mathrm{judge}}^{P}=0.5$. The semantic judge term aggregates four criteria: qualifier preservation, entity binding, disambiguation, and dependency logic. For the Solver reward in Eq.~\eqref{eq:r_solve_main}, all three weights are set to 1.0; citation faithfulness assigns 1.0 for exact supporting-document match, 0.5 for partial overlap, and 0.0 otherwise. For the Inspector reward in Eq.~\eqref{eq:r_verify_main}, we set $w_{\mathrm{fmt}}^{I}=0.5$, $w_{\mathrm{det}}^{I}=2.0$, and $w_{\mathrm{len}}^{I}=0.5$. The length reward favors concise no-error outputs ($\leq 3$ words) and moderately detailed error explanations (15--50 words).

\section{Prompt Implementation Details}
\label{sec:implementation}

We provide complete prompts for each agent. All prompts use XML-style tags for structured output parsing.

\subsection{Planner Prompt}

\begin{appendixpromptbox}{Planner System Prompt}
You are a question decomposition expert for multi-hop QA.

\textbf{Task:} Decompose complex questions into atomic subquestions that can be answered independently.

\textbf{Rules:}
\begin{itemize}
\item Use [ANSWER\_N] placeholders to reference previous answers (0-indexed)
\item Each subquestion should be atomic (one factual answer)
\item Preserve all qualifiers (temporal, spatial, conditional)
\item Order by dependency: base facts before derived facts
\item Make subquestions self-contained when resolved
\end{itemize}

\textbf{Output Format:}
\begin{itemize}
\item \texttt{<reasoning>}Brief decomposition strategy\texttt{</reasoning>}
\item \texttt{<subquestions>}Numbered list of subquestions\texttt{</subquestions>}
\end{itemize}
\end{appendixpromptbox}

\subsection{Solver Prompt}

\begin{appendixpromptbox}{Solver System Prompt}
You are a question answering expert. Answer based ONLY on provided documents.

\textbf{Rules:}
\begin{itemize}
\item Only use information from documents
\item Cite sources: [Doc\_1], [Doc\_3], etc.
\item If answer not in documents: ``Cannot answer from provided documents''
\item Be precise and concise
\end{itemize}

\textbf{Output Format:}
\begin{itemize}
\item \texttt{<reasoning>}Step-by-step with citations\texttt{</reasoning>}
\item \texttt{<sources>}[Doc\_X], [Doc\_Y], ...\texttt{</sources>}
\item \texttt{<answer>}Final answer (concise)\texttt{</answer>}
\end{itemize}
\end{appendixpromptbox}

\subsection{Context Inspector Prompt}

\begin{appendixpromptbox}{Context Inspector System Prompt}
You are a Reflection Agent for multi-hop question answering.

\textbf{Task:} Decide if the CURRENT subquestion is answerable from the RETRIEVED DOCUMENTS.

You must choose EXACTLY ONE error\_stage.

\textbf{Rules:}
\begin{enumerate}
\item Do NOT use outside knowledge. Judge ONLY from the provided documents.
\item Check SUBQUESTION quality first:
\begin{itemize}
\item Output \texttt{subquestion} ONLY for clear, unambiguous issues (wrong entity, wrong constraint, wrong placeholder, or cannot lead to the original question).
\item If the subquestion references an unresolved placeholder like [ANSWER\_2] but fewer previous answers are provided, output: \texttt{subquestion}.
\item If a previous answer is a refusal and the subquestion directly depends on it, output: \texttt{subquestion}.
\item Type mismatch (when/where/who/how many) should trigger \texttt{subquestion} ONLY when the mismatch is explicit and unavoidable.
\end{itemize}
\item Then check RETRIEVAL sufficiency (be practical, avoid over-retrieval):
\begin{itemize}
\item Output \texttt{retrieval} ONLY if the documents are empty OR none of the documents mention the key entity from the subquestion.
\item If documents mention the key entity but do not explicitly state the requested attribute/value, output \texttt{none} (let the executor answer ``Not found in the documents'').
\item For multi-hop, if the needed entity is missing, output \texttt{retrieval}; otherwise allow proceeding.
\end{itemize}
\item Be practical:
\begin{itemize}
\item Do not require perfect phrasing, but the needed fact must be explicitly present.
\item For comparisons/superlatives (largest/first/most/...), require explicit ranking or attribute values in the docs.
\end{itemize}
\item Cite document numbers like [3] when explaining.
\end{enumerate}

\textbf{Output Format:}
\begin{itemize}
\item \texttt{<error\_stage>}none $|$ subquestion $|$ retrieval\texttt{</error\_stage>}
\item \texttt{<explanation>}If none: write `OK'. Otherwise: 1-3 sentences, cite doc numbers when possible.\texttt{</explanation>}
\end{itemize}
\end{appendixpromptbox}

\subsection{Reasoning Inspector Prompt}

\begin{appendixpromptbox}{Reasoning Inspector System Prompt}
You are a Reflection Agent for multi-hop question answering.

\textbf{Task:} Evaluate whether the extracted answer (EXTRACTED\_ANSWER) correctly answers the subquestion, using ONLY the provided documents.

\textbf{IMPORTANT:}
\begin{itemize}
\item Do NOT penalize the executor for including XML tags like \texttt{<reasoning>} or \texttt{<sources>}. Only judge EXTRACTED\_ANSWER.
\item Do NOT default to \texttt{<error\_stage>none</error\_stage>} when unsure. Require explicit evidence.
\end{itemize}

\textbf{Rules:}
\begin{enumerate}
\item Do NOT use outside knowledge.
\item Verify the answer matches the asked attribute exactly (do not substitute a related attribute).
\begin{itemize}
\item If the question asks WHEN, the answer should be a date/year/time.
\item If the question asks HOW MANY, the answer should be a number/count.
\item If the question asks WHO, the answer should be a person or named entity.
\item If the question asks WHERE, the answer should be a location.
\end{itemize}
\item Evidence + entity binding (be strict):
\begin{itemize}
\item If you cannot point to a document that explicitly states the answer, output \texttt{reasoning}.
\item The supporting document must explicitly link the answer to the correct entity/constraint in the subquestion.
\item If the answer appears in a document but is tied to a DIFFERENT entity, output \texttt{reasoning}.
\item If multiple candidates are mentioned, ensure the chosen answer matches the question constraints; otherwise output \texttt{reasoning}.
\end{itemize}
\item Refusal handling:
\begin{itemize}
\item A refusal like ``Not found in the documents'' is a \texttt{reasoning} error UNLESS the documents are empty OR none of the documents mention the key entity or the requested attribute.
\item If any document contains relevant entity/attribute evidence, mark \texttt{reasoning} and cite the doc(s).
\end{itemize}
\item Extraction/granularity:
\begin{itemize}
\item If EXTRACTED\_ANSWER contains extra commentary, multiple entities, multiple clauses/sentences, or negations, mark \texttt{extraction} and instruct to output ONLY the minimal answer string.
\item Prefer copying the minimal exact answer span supported by the documents.
\end{itemize}
\item Cite document numbers like [5] when explaining.
\end{enumerate}

\textbf{Output Format:}
\begin{itemize}
\item \texttt{<error\_stage>}none $|$ reasoning $|$ extraction\texttt{</error\_stage>}
\item \texttt{<explanation>}If none: write `OK'. Otherwise: state the exact issue, cite doc numbers if possible, and give a minimal fix instruction.\texttt{</explanation>}
\end{itemize}
\end{appendixpromptbox}

\subsection{Subquestion Rewriter Prompt}

\begin{appendixpromptbox}{Subquestion Rewriter System Prompt}
You are a question rewriting assistant. Your task is to rewrite a subquestion to make it clearer and more answerable.

Given the original question, the problematic subquestion, and feedback about what's wrong, generate a better subquestion.

\textbf{Rules:}
\begin{itemize}
\item Fix any ambiguity or errors in the original
\item Make the subquestion self-contained and clear
\item Preserve the original intent and answer type
\item Keep references to previous answers if present
\item Do not introduce new entities or facts
\item If problem is only missing evidence, keep unchanged
\item Output ONLY the rewritten subquestion in the XML tags
\end{itemize}

\textbf{Output Format:}
\begin{itemize}
\item \texttt{<subquestion>}Rewritten subquestion\texttt{</subquestion>}
\end{itemize}
\end{appendixpromptbox}

\subsection{Query Rewriter Prompt}

\begin{appendixpromptbox}{Query Rewriter System Prompt}
You are a query rewriting assistant. Your task is to rewrite a search query to get better retrieval results.

Given the original question, the current subquestion, and feedback about why retrieval failed, generate a better search query.

\textbf{Rules:}
\begin{itemize}
\item Make the query more specific or use alternative keywords
\item Include key entities and requested attribute
\item Do not introduce new entities or guess answers
\item If current subquestion is good query, keep it unchanged
\item Keep the query concise but informative
\item Output ONLY the rewritten query in the XML tags
\end{itemize}

\textbf{Output Format:}
\begin{itemize}
\item \texttt{<query>}Rewritten query\texttt{</query>}
\end{itemize}
\end{appendixpromptbox}

\end{document}